\documentclass[letterpaper]{article} % DO NOT CHANGE THIS
\usepackage{aaai24}  % DO NOT CHANGE THIS
\usepackage{times}  % DO NOT CHANGE THIS
\usepackage{helvet}  % DO NOT CHANGE THIS
\usepackage{courier}  % DO NOT CHANGE THIS
\usepackage[hyphens]{url}  % DO NOT CHANGE THIS
\usepackage{graphicx} % DO NOT CHANGE THIS
\usepackage{natbib}  % DO NOT CHANGE THIS AND DO NOT ADD ANY OPTIONS TO IT
\usepackage{caption} % DO NOT CHANGE THIS AND DO NOT ADD ANY OPTIONS TO IT
\usepackage{algorithm}

\usepackage{amsmath}

\newtheorem{definition}{Definition}
\newtheorem{proposition}{Proposition}
\DeclareMathOperator*{\divop}{div}

\usepackage{newfloat}
\usepackage{listings}
\DeclareCaptionStyle{ruled}{labelfont=normalfont,labelsep=colon,strut=off} % DO NOT CHANGE THIS
\floatstyle{ruled}
\newfloat{listing}{tb}{lst}{}
\floatname{listing}{Listing}
\usepackage{amsfonts}
\usepackage{amsmath}
\usepackage{amssymb}
\usepackage[hyphens]{url}
\usepackage{url}
\usepackage{xcolor}
\usepackage{cleveref}
\usepackage[subrefformat=parens,labelformat=parens]{subfig}
\usepackage{threeparttable,booktabs}
\usepackage{makecell}
\usepackage{multirow}
\usepackage[]{algpseudocode}                               % algorithm package
\algrenewcommand\textproc{\texttt}
\makeatletter\let\float@addtolists\relax\makeatother
\usepackage{algorithm}
\usepackage{pgfplots}
\usepackage{pgfplotstable}
\pgfplotsset{compat=newest}
\usepackage{courier}
\usepackage[inline]{enumitem}
\usepackage{tabularx}
\usepackage{soul}
\usepackage{bm}
\usepackage{pifont}                                        % add circle
\usepackage{bbding}                                        % Checkmark
\usepackage[super]{nth}

\newcommand{\minisection}[1]{\noindent{\textbf{#1}}}

\renewcommand{\vec}[1]{\boldsymbol{#1}}                % re-define vec command

\DeclareMathOperator*{\argmax}{argmax}

\newtheorem{remark}{Remark}

\graphicspath{{./figs/}{../}}

\title{
    $p$-Laplacian Adaptation for Generative Pre-trained Vision-Language Models 
}
\author {
    Haoyuan Wu\equalcontrib,   \quad
    Xinyun Zhang\equalcontrib, \quad
    Peng Xu,                   \quad
    Peiyu Liao,                \quad
    Xufeng Yao,                \quad
    Bei Yu
}
\affiliations {
    Department of Computer Science \& Engineering, The Chinese University of Hong Kong\\
    \texttt{\small wuhyhowell@gmail.com, xyzhang21@cse.cuhk.edu.hk, pxu22@cse.cuhk.edu.hk,} \\ 
    \texttt{\small enzoliao95@gmail.com, yxf12fdu@gmail.com, byu@cse.cuhk.edu.hk}
}

\begin{document}

\maketitle
%%%%%%%%% ABSTRACT
\begin{abstract}
% Finetuning PLMs
% Efficient solution: adapter
% Adapter does not consider the special arch of attention
% Attention adapter: (1) graph message passing (2) heterophilic graphs and homophilic graphs (cross attention and self-attention) (3) p-laplacian adapter
% Exp: tasks and benchmarks
Vision-Language models (VLMs) pre-trained on large corpora have demonstrated notable success across a range of downstream tasks. In light of the rapidly increasing size of pre-trained VLMs, parameter-efficient transfer learning (PETL) has garnered attention as a viable alternative to full fine-tuning. One such approach is the adapter, which introduces a few trainable parameters into the pre-trained models while preserving the original parameters during adaptation.
In this paper, we present a novel modeling framework that recasts adapter tuning after attention as a graph message passing process on attention graphs, where the projected query and value features and attention matrix constitute the node features and the graph adjacency matrix, respectively. Within this framework, tuning adapters in VLMs necessitates handling heterophilic graphs, owing to the disparity between the projected query and value space.
To address this challenge, we propose a new adapter architecture, $p$-adapter, which employs $p$-Laplacian message passing in Graph Neural Networks (GNNs). Specifically, the attention weights are re-normalized based on the features, and the features are then aggregated using the calibrated attention matrix, enabling the dynamic exploitation of information with varying frequencies in the heterophilic attention graphs.
We conduct extensive experiments on different pre-trained VLMs and multi-modal tasks, including visual question answering, visual entailment, and image captioning. The experimental results validate our method's significant superiority over other PETL methods. Our code is available at \url{https://github.com/wuhy68/p-Adapter/}
\end{abstract}
% Vision-Language models (VLMs) pre-trained on massive corpus have shown significant competence on many downstream tasks. 
% Instead of full fine-tuning, parameter-efficient transfer learning is attracting more attention due to the rapidly increasing size of pre-trained models. 
% A representative approach is adapter, which inserts a few trainable parameters into the pre-trained models and keeps the original parameters intact during adaptation. 
% In this paper, we first propose a new modeling framework which re-frames adapter tuning after attention as message passing process in graph neural networks (GNNs), where the query and value features and the attention matrix are the nodes and adjacency matrix, respectively. Within this framework, the adaptation of VLMs requires handling heterophilic graphs due the the gap between the query space and the value space and the gap between the text and image modalities for cross-attention. 
% To tackle this challenge, we further propose a new adapter architecture, $p$-adapter, based on $p$-Laplacian message passing in GNNs. Specifically, the adjacency matrices are re-normalized according to the node features and features are aggregated using the calibrated attention matrix, to dynamically exploit the information with different frequencies in the heterophic graphs.
\section{Introduction}
% 1. PVLMs and downstream VL tasks
Recently, pre-trained language models (PLMs) \cite{BERT,gpt3,liu2019roberta,clark2020electra,2020t5,lewis-etal-2020-bart} have demonstrated significant success within the Natural Language Processing (NLP) community. 
By leveraging massive amounts of unlabeled data during training, PLMs can learn highly performant and generalizable representations, leading to improvements on various downstream tasks. Similarly, researchers have successfully applied massive pre-training techniques to generative vision-language models (VLMs) \cite{li2022blip,li2022mplug,cho2021unifying}. Using a sequence-to-sequence approach, generative VLMs can align cross-modal representations, which benefits multi-modal downstream tasks such as image captioning \cite{lin2014microsoft,sidorov2020textcaps,gurari2020captioning}, visual question answering (VQA) \cite{chen2020uniter,goyal2017making}, etc.
% Training on massive unlabeled corpus enables PLMs to learn performant and well-generalizable representations, leading to improvements on various downstream natural language understanding and generation tasks. Similarly, researchers have duplicated the success of massive pre-training on generative vision-language models (VLMs) \cite{li2022blip,li2022mplug,cho2021unifying}. Through a sequence-to-sequence manner, generative VLMs are able to align cross-modal representations, benefitting multi-modal downstream tasks, such as image captioning \cite{karpathy2015deep,lin2014microsoft,sidorov2020textcaps,gurari2020captioning}, visual question answering (VQA) \cite{chen2020uniter,goyal2017making}, and etc.

% 2. Parameter-efficient transfer learning
To effectively transfer the knowledge gained by pre-trained VLMs to downstream tasks, fine-tuning \cite{BERT,howard-ruder-2018-universal} has become the de facto paradigm, whereby all parameters of the model are tuned for each downstream task. However, as model sizes continue to grow rapidly, fine-tuning is increasingly affected by the parameter-efficiency issue \cite{houlsby2019parameter}. To address this challenge, \cite{sung2022vl,houlsby2019parameter} have proposed a solution that involves the use of adapters, which are small, learnable modules that can be inserted into the transformer blocks. By only tuning the adapters added to the model for each downstream task while keeping the original pre-trained model fixed, this approach achieves high parameter-efficiency and has demonstrated promising results on various downstream tasks.
% To effectively transfer the knowledge learned in pre-trained VLMs to downstream tasks, fine-tuning \cite{BERT,howard-ruder-2018-universal} has been a de facto paradigm, which tunes all the parameters for each downstream task. However, as the model size grows rapidly, fine-tuning suffers severely from the parameter-efficiency issue \cite{houlsby2019parameter}. To tackle this challenge, \cite{sung2022vl,houlsby2019parameter} proposed adapter, a small learnable module inserted into the transformer blocks. For each downstream task, only the added adapters are tuned while the original pre-trained model is fixed, which achieves great parameter-efficiency.

% 3. Formulating the attention adapter as a message passing process
% 4. Existence of heterophilic graphs
This paper introduces a novel modeling framework for adapter tuning coupled with attention. Specifically, we reformulate tuning adapters after attention to the spectral message passing process in GNNs on the attention graphs, wherein nodes and the edge weights are the projected query and value features and the attention weights, respectively. The attention graphs are bipartite, and each edge connects a feature in the projected query space and one in the projected value space. The discrepancy between the two feature spaces renders the attention graphs heterophilic graphs, in which the neighborhood nodes have distinct features \cite{pmlr-v162-fu22e,zhu2020beyond,tang2022rethinking}. Within this framework, the standard adapter \cite{sung2022vl,houlsby2019parameter} tuning process becomes a spectral graph convolution with the adjacency matrix serving as the aggregation matrix on the attention graphs. However, this graph convolution is impractical for handling heterophilic graphs \cite{pmlr-v162-fu22e}.
% In this paper, we propose a new modeling framework for adapter tuning coupled with attention. Specifically, we reframe tuning the adapters appended after attention to message passing process in GNNs, in which the nodes and adjacency matrix are the features projected into query and value spaces and the attention weights, respectively.
% The graphs are all bipartite, and each edge connects a feature in query space and a feature in value space, leading to heterophilic graphs. Besides, the cross-attention modules in generative VLMs also connects cross-modal features, and this modality gap even deepens the heterophily in the graphs. 
% Within this framework, vanilla adapter tuning becomes spectral graph convolution with adjacency matrix as aggregation matrix \cite{sung2022vl,houlsby2019parameter}, which is infeasible to handle the heterophic graphs \cite{pmlr-v162-fu22e}.

% 5. p-laplacian adapter 
To mitigate this heterophilic issue, we propose a new adapter module, $p$-adapter. Same as the vanilla adapter, $p$-adapter only has a small number of learnable parameters. What distinguishes $p$-adapter is to incorporate node features to renormalize the attention weights in pre-trained VLMs, which are further used for aggregating the features, inspired by $p$-Laplacian message passing \cite{pmlr-v162-fu22e}. \cite{pmlr-v162-fu22e} proves that by carefully choosing a renormalization factor, this renormalization and aggregation process can dynamically exploit the high- and low-frequency information in the graphs, thus achieving significant performance on heterophilic graphs. Therefore, with the $p$-Laplacian calibrated attention weights, tuning $p$-adapters in pre-trained VLMs can effectively capture the information with different frequencies in the heterophilic bipartite attention graphs. Additionally, the renormalisation intensity depends on the renormalization factor $p$. Unlike $p$-Laplacian message passing \cite{pmlr-v162-fu22e} with a fixed $p$, we adopt a layer-wise learnable strategy for determining $p$, thus rendering $p$-adapters more flexible to handle the attention graphs with different spectral properties.

% 6. Exp results
We conduct extensive experiments to validate the effectiveness of our proposed method. Specifically, we test $p$-adapter on six benchmarks related to three vision-language tasks: visual question answering \cite{goyal2017making, gurari2018vizwiz}, visual entailment \cite{song2022clip}, and image captioning \cite{lin2014microsoft,sidorov2020textcaps,gurari2020captioning}, using two generative pre-trained VLMs: BLIP \cite{li2022blip} and mPLUG \cite{li2022mplug}. Experimental results show our method's significant superiority over other PETL methods.

% % summary
% We summarize our contributions as follows:
% \begin{itemize}
% \item We propose a new modeling framework for adapter tuning in pre-trained VLMs, which proves that tuning adapters after attention modules equals to message passing process in GNNs. 

% \item To handle the heterophilic graphs, we propose $p$-adapter, which using $p$-Laplacian matrix to calibrate the attention weights.

% \item Extensive results demonstrate that our method can outperform other PETL methods on various downstream vision-language tasks.
% \end{itemize}

\section{Related Works}
% 1. VL models
\minisection{Vision-language models (VLMs).} 
VLMs integrate the text and image features into an aligned representation space. Single-encoder models \cite{su2019vl,li2020oscar,kim2021vilt,li2021align} train a unified cross-modal encoder using masked language/image modeling. Dual-encoder models \cite{clip,jia2021scaling} leverage two encoders for vision and language separately and include contrastive learning to align their representations. 
Encoder-decoder models\cite{li2022mplug,li2022blip,wang2022ofa,alayrac2022flamingo}, a.k.a. generative models, use a sequence-to-sequence approach for output, which endows VLMs with significant flexibility and paves the way for transferring to various generative downstream tasks, such as image captioning. In this work, we focus on the efficient adaptation of generative pre-trained VLMs.  

% 2. paremeter transfer learning
\minisection{Parameter-efficient transfer learning (PETL).}
Fine-tuning \cite{BERT} has long been the default paradigm for transferring knowledge from pre-trained models to downstream tasks. However, the rapid increase in the size of pre-trained models has led to severe parameter-efficiency \cite{houlsby2019parameter} issues for full fine-tuning. To bridge this gap, many PETL methods have been proposed \cite{houlsby2019parameter,hu2021lora,Li2021PrefixTuningOC}. Adapter-based methods \cite{houlsby2019parameter,hu2021lora,sung2022lst} introduce small learnable modules, called adapters, into pre-trained models and only fine-tune the inserted parameters during adaptation.
Another trend for PETL is prefix-/prompt-tuning \cite{lester-etal-2021-power,Li2021PrefixTuningOC}, which prepends several learnable token vectors to the keys and values in attention modules or to the input sequence directly. In this paper, we propose a new adapter architecture to handle the heterophily issue within our proposed modeling framework.          

% 3. graph message passing
\minisection{Graph Neural Networks (GNNs).}
GNNs \cite{kipf2017semisupervised, pmlr-v97-wu19e,ICLR2018-GAN,abu2018watch,pmlr-v162-fu22e} are neural networks operate on graph-structured data. Early GNNs are motivated from the spectral perspective, such as ChebNet~\cite{defferrard2016convolutional}. Graph Convolution Networks (GCNs)\cite{kipf2017semisupervised, pmlr-v97-wu19e}  further simplify ChebNet and reveal the message-passing mechanism of modern GNNs. $p$-GNN \cite{pmlr-v162-fu22e} proposes $p$-Laplacian graph message passing to handle heterophilic graphs. In this paper, we propose a new modeling framework for adapter tuning after attention, which reformulates it into a spectral graph message passing on the attention graphs and identifies the heterophilic issue therein. 

\iffalse
By utilizing graph message passing, GNNs can propagate the node features via the graph topology and capture the knowledge embedded within the graphs. Early GNNs are motivated from the spectral perspective, such as ChebNet~\cite{defferrard2016convolutional} that approximates the graph laplacian operators using recursive summation to reduce computation cost. An important variant of GNNs is Graph Convolution Networks (GCNs) 

% , chen2018fastgcn, pmlr-v80-xu18c, pmlr-v162-fu22e
\cite{kipf2017semisupervised, pmlr-v97-wu19e, chen2018fastgcn, pmlr-v80-xu18c, pmlr-v162-fu22e}, which simplify ChebNet~\cite{defferrard2016convolutional} by using its first order approximation and reveal the message-passing mechanism of modern GNNs. SGC \cite{pmlr-v97-wu19e} further simplifies the GCN architecture by removing non-linearities. $p$-GNN \cite{pmlr-v162-fu22e} proposes $p$-Laplacian graph message passing to handle heterophilic graphs. Graph Attention Networks (GATs) \cite{ICLR2018-GAN, abu2018watch} represent another category of GNNs that leverage attention mechanisms to aggregate node features instead of using spectral convolution during message passing. In this paper, we propose a new modeling framework for adapter tuning after attention, which reformulates it into spectral graph message passing on the attention graphs and identifies the heterophilic issue therein.   
\fi

\section{Method}
\label{sec:method}
This section begins with a brief review of the preliminaries, encompassing graph message passing, adapters, and attention mechanism. We then present an approach to model adapter tuning after attention to graph message passing and unveil the heterophilic issue therein. Finally, we propose a new adapter architecture, $p$-adapter, to address this issue.
\subsection{Preliminaries}
% 1. Perliminary: (1) Graph message passing (2) Adapter architecture

\minisection{Graph message passing.} Let $\mathcal{G}=(\mathcal{V}, \mathcal{E})$ be an undirected graph with node set $\mathcal{V}$ and edge set $\mathcal{E}$. Denote the node features $\vec{X} \in \mathbb{R}^{N \times d}$ and adjacency matrix $\vec{A} \in \mathbb{R}^{N \times N}$, where $N$ is the number of nodes, $d$ is node feature dimension, and $A_{ij}$ represents the edge weights between the $i$-th and the $j$-th node. The Laplacian matrix is defined as follows:
\begin{equation}
    \vec{L} = \vec{D} - \vec{A},
\end{equation}
where $\vec{D}$ is the degree matrix with diagonal entries $D_{ii}=\sum _{j}A_{ij}$ for $1\leq i\leq N$.
To propagate the node features and exploit the graph information, the spectral graph message-passing process can be defined as:
\begin{equation}
    \vec{X}^\prime = \sigma(\vec{C}\vec{X}\vec{W}),
    \label{eq:GCN-encoding}
\end{equation}
where $\vec{X}^\prime$ is the node embeddings after propagation, $\vec{W} \in \mathbb{R}^{d \times d}$ is a learnable weight, $\sigma(\cdot)$ is a non-linear function, e.g., $\mathrm{ReLU}(\cdot)$ \cite{Agarap2018DeepLU}, and $\vec{C}$ is the aggregation matrix. The choice of $\vec{C}$ depends on the spectral properties one wishes to obtain. For instance, original GCN \cite{kipf2017semisupervised} adopts $\vec{C}=\vec{D}^{-1/2}(2\vec{I}-\vec{L})\vec{D}^{-1/2}$, which serves as a low-pass filter \cite{pmlr-v97-wu19e} on the spectral domain. To overcome the over-smoothing issue \cite{chien2021adaptive}, \cite{pmlr-v162-fu22e,pmlr-v80-xu18c} propose to add residual connections to previous layers. 

\minisection{Adapter.} \cite{houlsby2019parameter, sung2022vl} propose inserting adapters into pre-trained models and only tuning the added parameters for better parameter efficiency. An adapter is a small learnable module containing two matrices $\vec{W}_{\text{down}} \in \mathbb{R}^{l_1 \times l_2}$, $\vec{W}_{\text{up}} \in \mathbb{R}^{l_2 \times l_1}$ and a non-linear function $\sigma(\cdot)$, where $l_1$ and $l_2$ are the feature dimensions in pre-trained models and the hidden dimension in adapter (usually $l_2 < l_1$). Given a feature $\vec{U} \in \mathbb{R}^{N \times l_1}$ in the pre-trained model, the adapter encoding process can be represented as:
\begin{equation}
    \vec{U}^\prime = \sigma(\vec{U}\vec{W}_{\text{down}})\vec{W}_{\text{up}} + \vec{U}.
    \label{eq:adapter-encoding}
\end{equation}
\cite{houlsby2019parameter,sung2022vl} place two adapters right after the attention and the feed-forward network in the transformer block, respectively.  

\minisection{Attention} \cite{vaswani2017attention} has been the basic building block for foundation models \cite{li2022blip,gpt3,wang2022ofa}. Given query $\vec{Q} \in \mathbb{R}^{N_1 \times d_k}$, key $\vec{K} \in \mathbb{R}^{N_2 \times d_k}$ and value $\vec{V} \in \mathbb{R}^{N_2 \times d_v}$, attention aggregates the features by:
\begin{equation}
    \mathrm{Attn}(\vec{Q}, \vec{K}, \vec{V})=\vec{M}\vec{V},
    \label{eq:attn-cal}
\end{equation}
where 
\begin{equation}
\vec{M} = \mathrm{softmax}\left(\frac{\vec{Q}\vec{K}^{\top}}{\sqrt{d_k}}\right)
\label{eq:attn-weights}
\end{equation}
represents the attention weights, $N_1$ and $N_2$ are the number of query and key/value features, respectively. Multi-head attention further transforms the query, key and value onto multiple sub-spaces and calculates attention on each of them, which can be formulated as:
\begin{equation}
     \mathrm{MHA}(\vec{Q}, \vec{K}, \vec{V}) =\mathrm{Concat}(\mathrm{head}_1, \cdots, \mathrm{head}_n)\bm{W}_o, 
     \label{eq:attn-mha}
\end{equation}
where
\begin{equation}
    \mathrm{head}_i = \mathrm{Attn}(\vec{Q}\vec{W}_q^{i}, \vec{K}\vec{W}_k^{i}, \vec{V}\vec{W}_v^{i}),
    \label{eq:attn-mha-single}
\end{equation}
$\vec{W}_o \in \mathbb{R}^{d_v \times d_v}$, $n$ is the number of heads and $\vec{W}_q^{i}, \vec{W}_k^{i} \in \mathbb{R}^{d_k \times \frac{d_k}{n}} , \vec{W}_v^{i} \in \mathbb{R}^{d_v \times \frac{d_v}{n}}$ are the transformation matrices for the query, key and value in the $i$-th head, respectively.

\subsection{Modeling Adapter as Graph Message Passing}
% \begin{figure*}
%     \centering
%     % \hspace{-2cm}
%     \subfloat[]{ \includegraphics[height=2.5cm]{figs/attn-1.pdf} \label{fig:attn-1} } %\hspace{2cm}
%      \subfloat[]{ \includegraphics[height=1.8cm]{figs/attn-2.pdf}          \label{fig:attn-2} }  
%     %\hspace{2cm}
%     \subfloat[]{ \includegraphics[height=2.5cm]{figs/attn-3.pdf}          \label{figa:attn-3} }
%         \caption{(a):  (b): (c): }
%     \label{fig:attn}
% \end{figure*}

% \begin{figure}
%     \centering
%      \subfloat[]{ \input{figs/tsne_self.tex}          \label{fig:tsne-self} }  
%     %\hspace{2cm}
%     \subfloat[]{ \input{figs/tsne_cross.tex}          \label{fig:tsne-cross} }
%         \caption{(a):  (b): (c): }
%     \label{fig:attn}
% \end{figure}

\begin{figure}[tb!]
    \centering
    % \hspace{-0.4cm}
    \subfloat[Attn.]{\includegraphics[height=2.4cm]{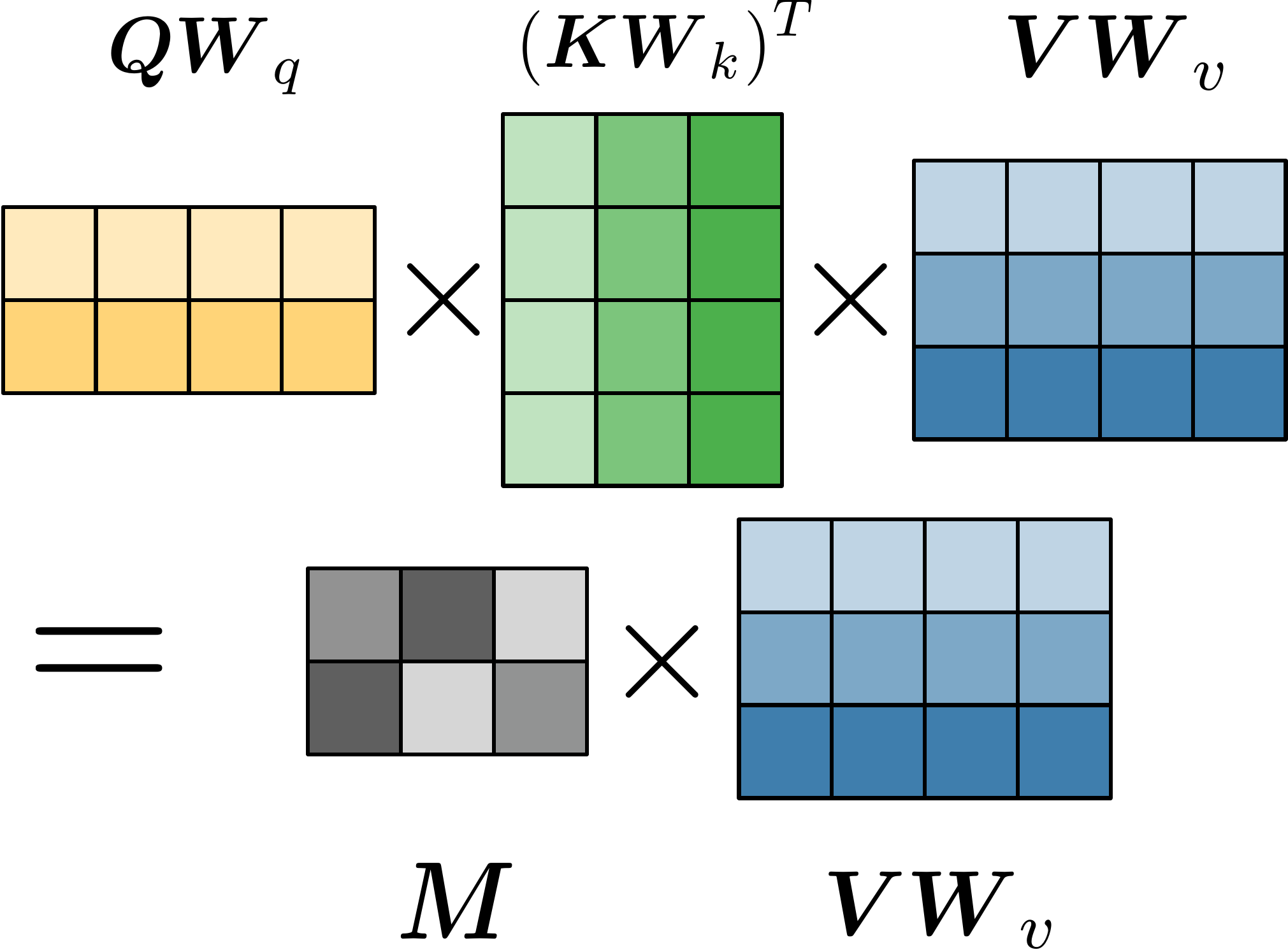} 
    \label{fig:attn-1}} %\hspace{2cm}
    \subfloat[Augmented Attn.]{\includegraphics[height=2.1cm]{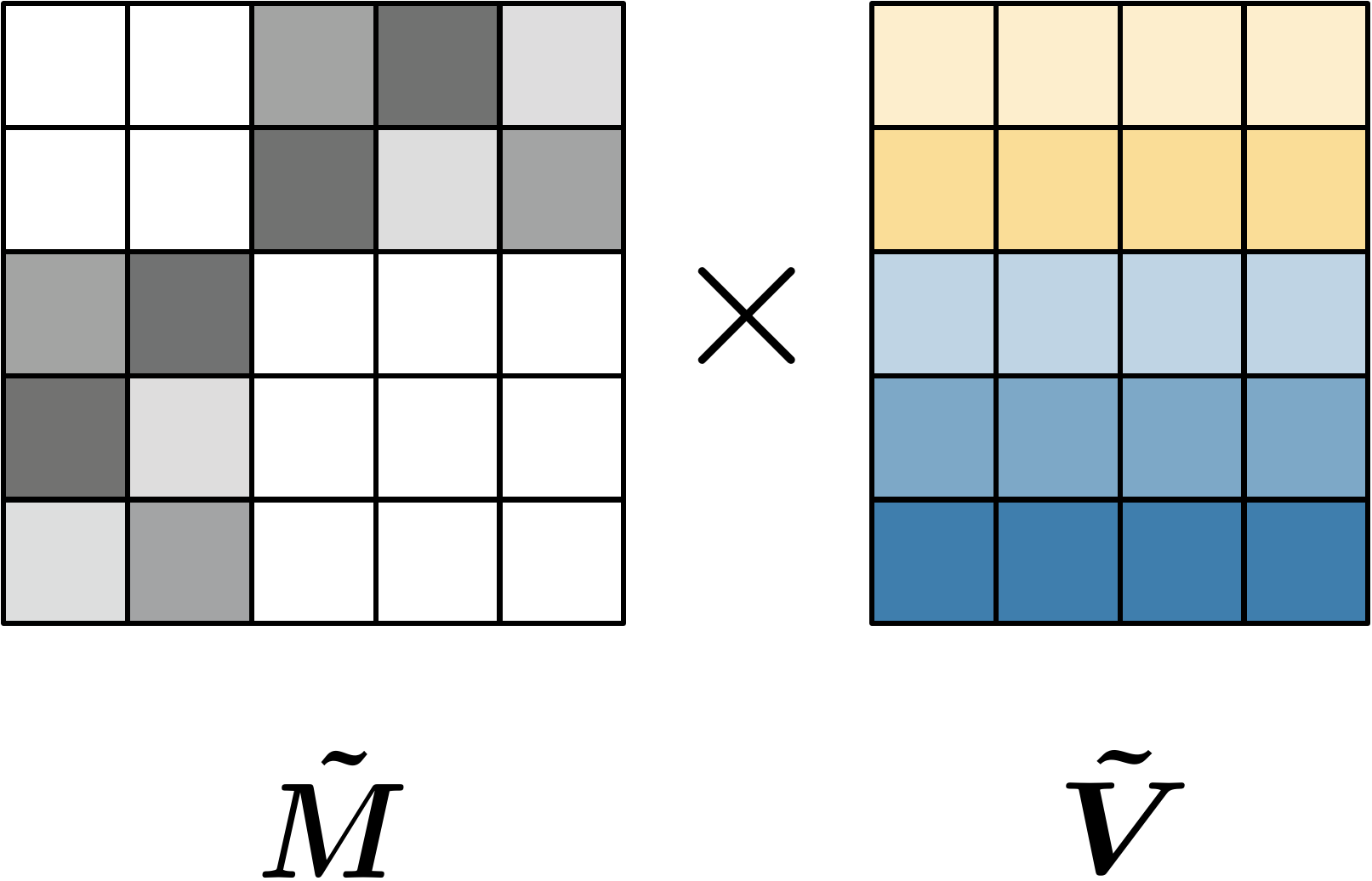}          
    \label{fig:attn-2}}  
    \subfloat[Attn. graph]{\includegraphics[height=2.4cm]{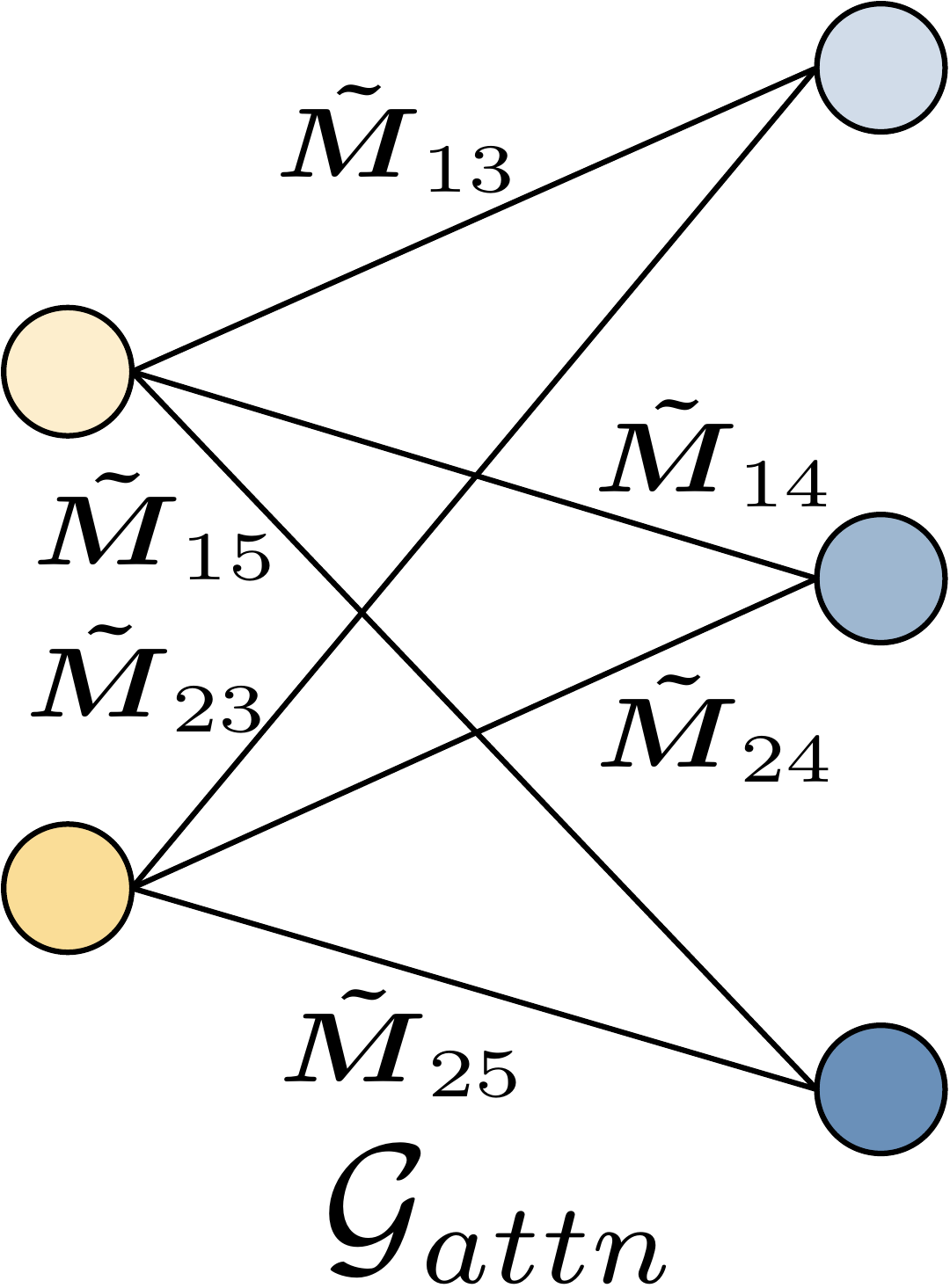}          
    \label{figa:attn-3}}
    \caption{Illustration of the generation of the bipartite attention graph $\mathcal{G}_{attn}$. For simplicity, we omit the scale and softmax functions in attention mechanism.}
    \label{fig:attn-graph}
\end{figure}

% (1) formulation (2) existence of homophilic and heterophilic graphs
In this section, we propose a new framework to model adapter \textit{after attention} as spectral graph message passing. To simplify the notation, we consider single-head attention. From \Cref{eq:adapter-encoding} and \Cref{eq:attn-cal}, we can formulate the features sequentially encoded by attention and adapter as:
\begin{equation}
\vec{U}^\prime=\sigma(\vec{M}\vec{V}\vec{W}_v\vec{W}_o\vec{W}_{\text{down}})\vec{W}_{\text{up}} + \vec{M}\vec{V}\vec{W}_v\vec{W}_o,
    \label{eq:attn-adapter-encoding}
\end{equation}
where $\vec{M} \in \mathbb{R}^{N_1 \times N_2}$ is the attention matrix computed by the transformed query $\vec{Q}\vec{W}_q$ and key $\vec{K}\vec{W}_k$ using \Cref{eq:attn-weights}. The key idea of this modeling framework is to construct an attention graph, where the attention weights and features are the edge weights and node embeddings, respectively. With this attention graph, we can transform the adapter encoding process in \Cref{eq:attn-adapter-encoding} into spectral graph message passing. 
% However, direct mapping is infeasible since $\vec{M}$ is not necessarily a \textbf{square} and \textbf{symmetric} matrix and thus can not be regarded as the adjacency matrix of a graph. 
However, direct mapping can not be applied since the adjacency matrix of an undirected graph must be \textbf{square} and \textbf{symmetric} and neither self-attention nor cross-attention satisfies this condition. For self-attention, the asymmetry arises from the distinct transform matrices of query and key spaces. For cross-attention, the attention matrix is not necessarily square if the number of query vectors does not match that of the key vectors.
To bridge this gap, we consider an augmented attention mechanism. Specifically, supposing both the transformed query and value share the same dimension, we define the augmented value feature $\Tilde{\vec{V}}$ which concatenates the transformed query and value and the augmented attention matrix $\Tilde{\vec{M}}$ as
\begin{equation}
    \Tilde{\vec{V}}=\begin{bmatrix}
    \vec{Q}\vec{W}_q \\ \vec{V}\vec{W}_v 
    \end{bmatrix}
    ,\quad \Tilde{\vec{M}}=\begin{bmatrix}
    \vec{0} & \vec{M} \\ \vec{M}^{\top} & \vec{0}
    \end{bmatrix}.
    \label{eq:augmented-attention}
\end{equation}
The attention output $\vec{M}\vec{V}\vec{W}_v$ equals to the first $N_1$ features from the output of the augmented attention $\Tilde{\vec{M}}\Tilde{\vec{V}}$, as shown in \Cref{fig:attn-1} and \Cref{fig:attn-2}.
Defining the projected augmented value feature $\Hat{\vec{V}}=\Tilde{\vec{V}}\vec{W}_o$, with the augmented attention mechanism, we can further define the augmented adapter encoding process by:
\begin{equation}
     \Tilde{\vec{U}}^\prime=\sigma(\Tilde{\vec{M}}\Hat{\vec{V}}\vec{W}_{\text{down}})\vec{W}_{\text{up}} + \Tilde{\vec{M}}\Hat{\vec{V}}.
     \label{eq:aug-attn-adapter-encoding}
\end{equation}
Comparing \Cref{eq:attn-adapter-encoding} and \Cref{eq:aug-attn-adapter-encoding}, we can obtain that $  \vec{U}^\prime=\Tilde{\vec{U}}^\prime_
{:N_1,:}$. This indicates that the adapter encoding process and the augmented one are equal, by just taking the first $N_1$ elements from $\Tilde{\vec{U}}^\prime$, thus we transform the adapter encoding process into the augmented one. 
Since $\Tilde{\vec{M}}$ is a square and symmetric matrix, we can regard it as the adjacency matrix of the attention graph $\mathcal{G}_{attn}$, in which the nodes are features from $\Hat{\vec{V}}$, i.e.,  $\vec{Q}\vec{W}_q\vec{W}_o$ and $ \vec{V}\vec{W}_v\vec{W}_o$, and the edge weights are the corresponding attention weights computed by \Cref{eq:augmented-attention}. Note that during adaptation $\vec{W}_q$, $\vec{W}_v$ and $\vec{W}_o$ are all fixed, thus $\vec{Q}\vec{W}_q\vec{W}_o$ and $\vec{V}\vec{W}_v\vec{W}_o$ are linearly projected query and value space, respectively. Therefore, we can approximate the augmented adapter encoding process by spectral graph message passing process in \Cref{eq:GCN-encoding}, by setting $\vec{C}=\Tilde{\vec{M}}$ and $\vec{W}=\vec{W}_{\text{down}}$, 
considering the shortcut in adapter as a residual connection in GCN \cite{pmlr-v80-xu18c}, and regarding $\vec{W}_{\text{up}}$ as a linear mapping after each graph convolution layer. In other words, adapter tuning can be regarded as using a one-layer GCN with the adjacency matrix serving as the aggregation matrix on $\mathcal{G}_{attn}$. Through this, we can analyze adapter tuning from a graph perspective. Upon closer inspection of $\mathcal{G}_{attn}$, we observe it to be a bipartite graph with edges connecting nodes in the projected query and value spaces, as shown in \Cref{figa:attn-3}.
% heterophilic graph
\begin{figure}[tb!]
    \centering
    \hspace{-.4in}
    \subfloat[Self-attention]{\scalebox{0.8}{
\begin{tikzpicture}[]
    \definecolor{myquery}{HTML}{ffd791}
    \definecolor{myvalue}{HTML}{2073a2}
    \begin{axis}[
        xmin = -120, xmax = 120,
        ymin = -120, ymax = 120,
        width  = 0.6\linewidth,
        height = 0.45\linewidth,
        xticklabel style={font=\small}, 
        yticklabel style={font=\small},
        x tick label style={
            /pgf/number format/assume math mode, font=\scriptsize},
        y tick label style={
            /pgf/number format/assume math mode, font=\scriptsize},
        % ytick = {0.00, 0.04,..., 0.16},
        % yticklabels = {0.00,, 0.04, 0.08, 0.12, 0.16},
        % grid style = dashed,
        % xmajorgrids = true, ymajorgrids = true,
        legend style={
            draw=none,
            fill=none,
            at = {(rel axis cs:0.5, 1)},
            anchor = south,
            legend columns = 2,
            font=\small,
        }
    ]
    \addplot[myvalue, only marks, mark size=0.5] table {data/h.dat};    
    \addplot[myquery, only marks, mark size=0.5] table {data/g.dat};
    \legend{value, query}
    \end{axis}
\end{tikzpicture}
}   \label{fig:tsne-self}}
    \hspace{-.1in}
    \subfloat[Cross-attention]{\scalebox{0.8}{
\begin{tikzpicture}[]
    \definecolor{myquery}{HTML}{ffd791}
    \definecolor{myvalue}{HTML}{2073a2}
    \begin{axis}[
        xmin = -120, xmax = 120,
        ymin = -120, ymax = 120,
        width  = 0.6\linewidth,
        height = 0.45\linewidth,
        xticklabel style={font=\small}, 
        yticklabel style={font=\small},
        x tick label style={
            /pgf/number format/assume math mode, font=\scriptsize},
        y tick label style={
            /pgf/number format/assume math mode, font=\scriptsize},
        % ytick = {0.00, 0.04,..., 0.16},
        % yticklabels = {0.00, 0.04, 0.08, 0.12, 0.16},
        % grid style = dashed,
        % xmajorgrids = true, ymajorgrids = true,
        legend style={
            draw=none,
            fill=none,
            at = {(rel axis cs:0.5, 1)},
            anchor = south,
            legend columns = 2,
            font=\small,
        }
    ]
        
    \addplot[myvalue, only marks, mark size=0.5] table {data/v.dat};
    \addplot[myquery, only marks, mark size=0.5] table {data/q.dat};

    \legend{value,query}

    \end{axis}
\end{tikzpicture}
}  \label{fig:tsne-cross}}
    \hspace{-.4in}
    \caption{
        The t-SNE \cite{van2008visualizing} visualization of the features in the projected query and value space for self- and cross-attention.
        The VLM is BLIP$_\text{CapFilt-L}$ \cite{li2022blip} and data come from COCO Captions \cite{lin2014microsoft}.
    }
    \label{fig:tsne-vis}
\end{figure}
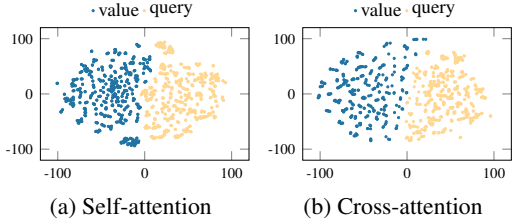

\begin{remark}
The attention graph $\mathcal{G}{attn}$ is a heterophilic graph in which connected nodes have dissimilar features. The visualization of the learned distribution of the projected query and value space is shown in \Cref{fig:tsne-vis}. We can observe that in both self- and cross-attention, the features from the two spaces differ significantly, forming two well-separated clusters. This indicates that each edge of $\mathcal{G}_{attn}$ connects two nodes from distinct feature spaces, underscoring its heterophily. 
\end{remark}
% GCN only suits for homophilic graphs
Most existing GCNs work under the homophilic assumption, which requires the labels or the features of the neighborhood nodes to be similar \cite{pmlr-v162-fu22e,zhu2020beyond}. When dealing with heterophilic graphs, the high-frequency information in the spectral domain will be infeasible for vanilla GCNs to exploit \cite{tang2022rethinking}, since they mostly perform as low-pass filters \cite{tang2022rethinking,pmlr-v97-wu19e,pmlr-v162-fu22e}. Therefore, the heterophilic nature of $\mathcal{G}_{attn}$ poses challenges for adapters, which is previously shown to be equivalent to vanilla spectral graph message passing. 

\subsection{$p$-Adapter}
% \begin{figure*}[tbp] 
%     \centering
%     % \hspace{-0.5cm}
%     \includegraphics[width=0.67\linewidth]{figs/p-adapter-architecture-2.pdf}
%     \caption{$p$-Adapter architecture.} 
%     \label{fig:p-adapter-arch}
% \end{figure*}
\begin{figure*}[tb!]
    \begin{minipage}{.67\linewidth}
        \centering
        \includegraphics[width=0.99\linewidth]{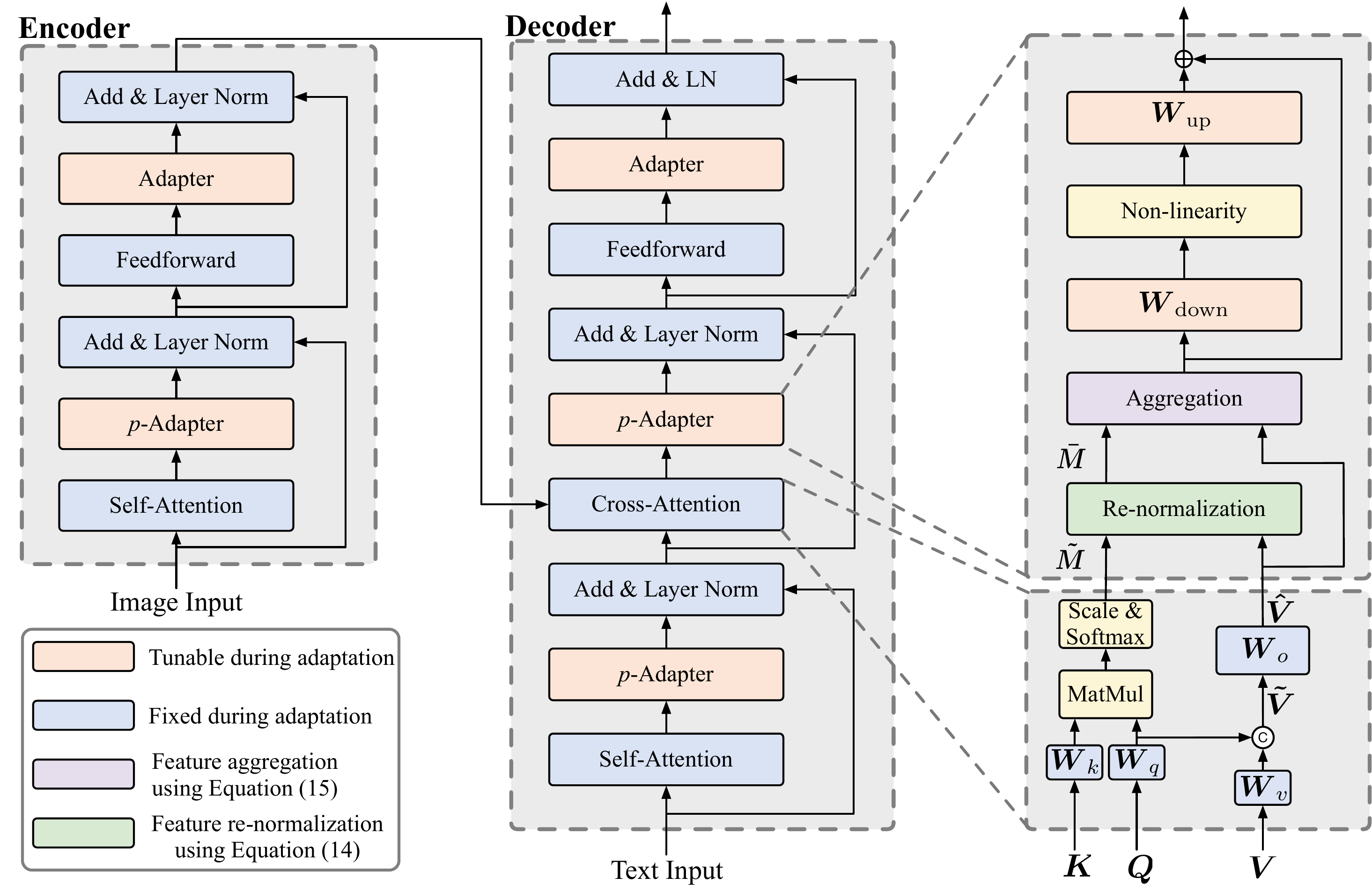} 
        \caption{$p$-Adapter architecture.}
        \label{fig:arch}
    \end{minipage}
    \hfill
    \begin{minipage}{.35\linewidth}
        \centering
        \includegraphics[width=0.82\linewidth]{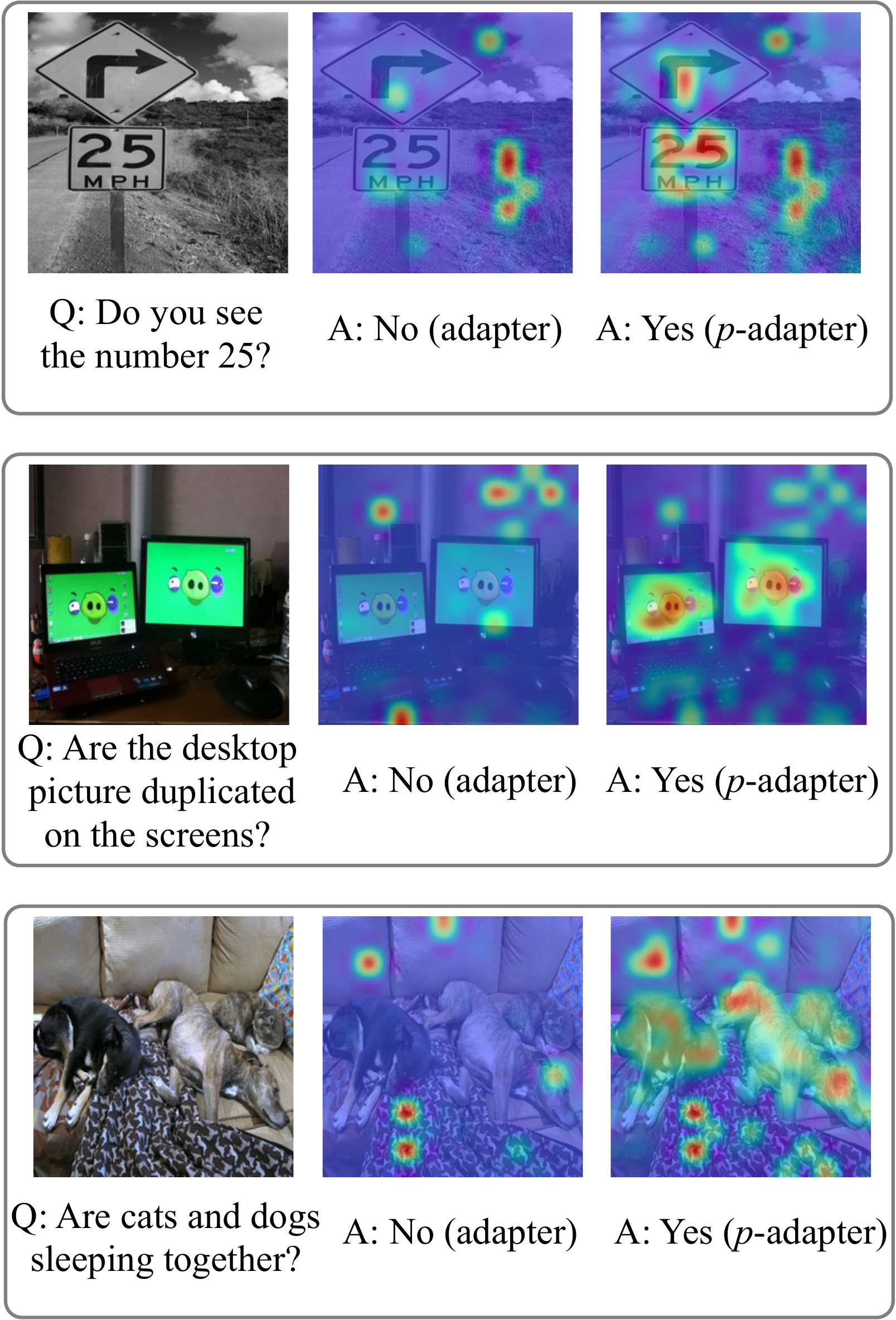} 
        \caption{Attention visualization.} 
    \label{fig:vis}
    \end{minipage}
\end{figure*}
% 1. p-Laplacian message passing 2. p-Laplacian adapter
To tackle the heterophilic issue in adapter learning, we propose a new adapter, $p$-adapter, inspired by $p$-Laplacian message passing \cite{pmlr-v162-fu22e}.

\minisection{$p$-Laplacian message passing} \cite{pmlr-v162-fu22e} is proposed for heterophilic graph learning. By denoting $\vec{\alpha}=\mathrm{diag}(\alpha_{1,1}, \cdots, \alpha_{N,N})$, $\vec{\beta}=\mathrm{diag}(\beta_{1,1},\cdots,\beta_{N,N})$, and two hyper-parameters $\mu,p \in \mathbb{R}$, one-layer $p$-Laplacian message passing can be defined as:
\begin{equation}
\vec{X}^\prime=\vec{\alpha}\vec{D}^{-1/2}\Bar{\vec{A}}\vec{D}^{-1/2}\vec{X}+\vec{\beta}\vec{X},
\label{eq:p-laplacian-forward}
\end{equation}
where $\Bar{\vec{A}}$ is the $p$-Laplacian normalized adjacency matrix with entries defined by:
\begin{equation}
    \Bar{A}_{i,j} = A_{i,j}\left\|\sqrt{\frac{A_{i,j}}{D_{i,i}}}\vec{X}_{i,:}-\sqrt{\frac{A_{i,j}}{D_{j,j}}}\vec{X}_{j,:}\right\|^{p-2},
    \label{eq:p-laplacian-norm}
\end{equation}
and for all $i,j \in [N]$ we have:
\begin{equation}
    \alpha_{i,i}=\left(\sum^{N}_{j=1}\frac{\Bar{A}_{i,j}}{D_{i,i}}+\frac{2\mu}{p}\right)^{-1},\quad \beta_{i,i}=\frac{2\mu}{p}\alpha_{i,i}.
    \label{eq:alpha-beta}
\end{equation}
The key idea of $p$-Laplacian message passing is to adopt the node features to re-normalize the adjacency matrix, as shown in \Cref{eq:p-laplacian-norm}. In other words, $p$-Laplacian message passing can adaptively learn the aggregation weights for different graph-structured data. The second term in \Cref{eq:p-laplacian-forward} $\vec{\beta}\vec{X}$ is a residual term to mitigate the oversmoothing issue \cite{chien2021adaptive}. The hyper-parameter $p$ controls the intensity of normalization, and different choices of $p$ lead to different spectral properties. When $p=2$, we impose no normalization and $p$-Laplacian message passing degenerates to vanilla GCN spectral message passing. When $p \in [1,2)$, \cite{pmlr-v162-fu22e} proves theoretically that $p$-Laplacian message passing works as low-pass filters for nodes with small gradient, i.e., nodes with similar neighbor nodes, and works as low-high-pass filters for nodes with large gradient, i.e., nodes with dissimilar neighbor nodes. This dynamic filtering property enables $p$-Laplacian message passing to be able to handle heterophilic graphs. In \cite{pmlr-v162-fu22e}, they adopt a fixed $p$ value and test different $p$ values for different tasks. 
See \Cref{appendix:C} for more details.

\minisection{$p$-Aadapter architecture.} To exploit the heterophilic attention graph $\mathcal{G}_{attn}$, the main idea of $p$-adapter is to leverage the node features to calibrate the weights of the attention/adjacency matrix, similar to $p$-Laplacian message passing \cite{pmlr-v162-fu22e}. The architecture of $p$-adapter is shown in \Cref{fig:arch}. The input to $p$-adapter is the intermediate result of attention. Suppose we consider the single-head case, the final output of attention should be $\vec{M}\vec{V}\vec{W}_o$, as shown in \Cref{eq:attn-cal} and \Cref{eq:attn-mha}. For $p$-adapter, we take the attention matrix $\vec{M}$ and the projected augmented value feature $\Hat{\vec{V}}$, as the output of attention. Note that this transformation does not alter any learned parameters in attention. Then, we augment the attention matrix to $\Tilde{\vec{M}}$, as shown in \Cref{eq:augmented-attention}. Following \Cref{eq:p-laplacian-norm}, we normalize the augmented attention matrix by:
\begin{equation}
    \Bar{M}_{i,j}= \Tilde{M}_{i,j}\left\|\sqrt{\frac{\Tilde{M}_{i,j}}{\Tilde{D}_{i,i}}}\Hat{\vec{V}}_{i,:}-\sqrt{\frac{\Tilde{M}_{i,j}}{\Tilde{D}_{j,j}}}\Hat{\vec{V}}_{j,:}\right\|^{p-2},
    \label{eq:p-adapter-normalization}
\end{equation}
where $\Tilde{\vec{D}}$ is the degree matrix of $\Tilde{\vec{M}}$. Then, we can obtain $\Tilde{\vec{\alpha}}$ and $\Tilde{\vec{\beta}}$ by replacing $\Bar{\vec{A}}$ and $\vec{D}$ with $\Bar{\vec{M}}$ and $\Tilde{\vec{D}}$ in \Cref{eq:alpha-beta}, respectively. Further, we can aggregate the features using the calibrated attention matrix $\Bar{\vec{M}}$ by 
\begin{equation}
    \Bar{\vec{U}}= \Tilde{\vec{\alpha}}\Tilde{\vec{D}}^{-1/2}\Bar{\vec{M}}\Tilde{\vec{D}}^{-1/2}\Hat{\vec{V}}+\Tilde{\vec{\beta}}\Hat{\vec{V}},
\end{equation}
similar to \Cref{eq:p-laplacian-forward}. With the aggregated feature $\Bar{\vec{U}}$, we encode it with the learnable adapter weights by:
\begin{equation}
\Bar{\vec{U}}^\prime=\sigma(\Bar{\vec{U}}\vec{W}_{\text{down}})\vec{W}_{\text{up}}+\Bar{\vec{U}}.
\end{equation}
The output of $p$-adapter is $\Bar{\vec{U}}^\prime_{:N_1,:}$, since we only collect the features aggregated on the query nodes in the attention graph $\mathcal{G}_{attn}$. By adopting the renormalization technique inspired by $p$-Laplacian message passing \cite{pmlr-v162-fu22e}, $p$-adapter can effectively handle the heterophilic issue in $\mathcal{G}_{attn}$ and lead to improvements on various downstream tasks. Moreover, unlike \cite{pmlr-v162-fu22e} using a fixed $p$ value, we adopt a layer-wise learnable strategy for determining value $p$. In addition, $p$-adapter is compatible with vanilla adapter \cite{houlsby2019parameter,sung2022vl}, since $p$-adapter is designed for adaptation after attention and we leave the adapter after feed-forward networks unchanged.

\section{Experiments}
\subsection{Tasks and Datasets}
We conduct experiments on six benchmarks related to three vision-language downstream tasks, i.e., visual question answering (VQA), visual entailment (VE) and image captioning. For VQA, we consider it as an answer generation problem, following \cite{cho2021unifying, li2022blip, li2021align}. We test our model on VQA2.0 \cite{goyal2017making} with the widely-used Karpathy split \cite{karpathy2015deep} and VizWizVQA \cite{gurari2018vizwiz}. The evaluation metric is accuracy.  
For VE, we follow the setting in \cite{song2022clip} and adopt SNLI-VE \cite{xie2019visual} as the evaluation benchmark, with accuracy as the metric.
For image captioning, we conduct extensive experiments on three benchmarks, i,e., COCO Captions \cite{lin2014microsoft} with Karpathy split \cite{karpathy2015deep}, TextCaps \cite{sidorov2020textcaps}, and VizWizCaps \cite{gurari2020captioning}. We adopt BLEU@4 \cite{papineni2002bleu} and CIDEr \cite{vedantam2015cider} as the evaluation metrics, same as \cite{li2022blip,yang2022prompt}. 
Please refer to \Cref{appendix:A} for more details.

\subsection{Implementation Details}
Our experiments are implemented in PyTorch \cite{paszke2019pytorch} and conducted on 8 Nvidia 3090 GPUs. We validate our method on two generative pre-trained VLMs, BLIP$_\text{CapFilt-L}$ \cite{li2022blip} and mPLUG$_\text{ViT-B}$ \cite{li2022mplug}. Specifically, we use the encoder of BLIP$_\text{CapFilt-L}$/mPLUG$_\text{ViT-B}$ to encode the image, and the decoder of BLIP$_\text{ViT-B}$/mPLUG$_\text{ViT-B}$ to generate the answers in an auto-regressive way. Following \cite{gu2022wukong, sung2022vl}, we freeze the encoder and only train the decoder for new tasks. We use AdamW \cite{loshchilov2017decoupled} optimizer with a weight decay of 0.05 and apply a linear scheduler. We take random image crops of resolution $224 \times 224$ as the input of the encoder, and also apply RandAugment \cite{cubuk2020randaugment} during the training, following \cite{li2022blip,sung2022vl}. We train the model for five and two epochs for VQA and VE, and image captioning, respectively. We sweep a wide range of learning rates over $\{1 \times 10^{-4}, 2 \times 10^{-4}, 5 \times 10^{-4}, 1 \times 10^{-3}\}$ for PETL methods, and use $2 \times 10^{-5}$ for full fine-tuning, same as \cite{sung2022vl}. 
Please refer to \Cref{appendix:B} for more details. 

\subsection{Comparison with transfer learning methods}

\begin{table*}[]
\centering
\setlength\tabcolsep{2pt}
\footnotesize
\begin{tabular*}{\hsize}{@{}@{\extracolsep{\fill}}cccccccccccc@{}}
\toprule
\multirow{3}{*}{Method} & \multirow{2}{*}{\makecell[c]{Updated \\ Params}} & VQA2.0 & VizWizVQA & SNLI\_VE & \multicolumn{2}{c}{COCOCaps} & \multicolumn{2}{c}{TextCaps} & \multicolumn{2}{c}{VizWizCaps} \\
 & & Karpathy test & test-dev & test-P & \multicolumn{2}{c}{Karpathy test} & \multicolumn{2}{c}{test-dev} & \multicolumn{2}{c}{test-dev} & \multirow{3}{*}[2.5ex]{\quad Avg. \quad}\\
& (\%) & Acc.(\%) & Acc.(\%) & Acc.(\%) & BLEU@4 & CIDEr & BLEU@4 & CIDEr & BLEU@4 & CIDEr \\
\midrule
\specialrule{0em}{1pt}{1pt}
\midrule
\multicolumn{12}{c}{BLIP$_\text{CapFilt-L}$}\\
\midrule
\multicolumn{1}{l}{Full fine-tuning} & 100.00 & \textbf{70.56} & \textbf{36.52} & \textbf{78.35} & \textbf{39.1} & \textbf{128.7} & \textbf{27.1} & \textbf{91.6} & \textbf{45.7} & \textbf{170.0} & \textbf{76.40} \\
\multicolumn{1}{l}{Prefix tuning} & 0.71 & 60.49 & 22.45 & 71.82 & 39.4 & 127.7 & 24.8 & 80.0 & 40.6 & 153.3 & 68.95\\
\multicolumn{1}{l}{LoRA} & 0.71 & 66.57 & 33.39 & 77.36 & 38.3 & 128.3 & 24.6 & 82.2 & 41.3 & 154.3 & 71.81 \\
\multicolumn{1}{l}{Adapter} & 6.39 & 69.53 & 35.37 & 78.85 & 38.9 & 128.8 & 25.4 & 86.7 & 43.3 & 160.5 & 74.15\\
\midrule
\multicolumn{1}{l}{$p$-Adapter (Ours)} & 6.39 & \textbf{70.39} & \textbf{37.16} & \textbf{79.40} & \textbf{40.4} & \textbf{130.9} & \textbf{26.1} & \textbf{87.0} & \textbf{44.5} & \textbf{164.1} & \textbf{75.54}\\
\midrule
\specialrule{0em}{1pt}{1pt}
\midrule
\multicolumn{12}{c}{mPLUG$_\text{ViT-B}$}\\
\midrule
\multicolumn{1}{l}{Full fine-tuning} & 100.00 & \textbf{70.91} & \textbf{59.79} & \textbf{78.72} & \textbf{40.4} & \textbf{134.8} & \textbf{23.6} & \textbf{74.0} & \textbf{42.1} & \textbf{157.5} & \textbf{75.76}\\
\multicolumn{1}{l}{Prefix tuning} & 0.71 & 60.95 & 47.42 & 72.11 & 39.8 & 133.5 & 18.8 & 51.9 & 35.5 & 135.6 & 66.18\\
\multicolumn{1}{l}{LoRA} & 0.71 & 66.67 & 52.49 & 75.29 & 39.4 & 129.4 & 21.0 & 64.4 & 39.5 & 146.0 & 70.46 \\
\multicolumn{1}{l}{Adapter} & 6.39 & 70.65 & 56.50 & 78.56 & 40.3 & 134.7 & 22.9 & 71.5 & 41.9 & 155.6 & 74.73 \\
\midrule
\multicolumn{1}{l}{$p$-Adapter (Ours)} & 6.39 & \textbf{71.36} & \textbf{58.08} & \textbf{79.26} & \textbf{40.4} & \textbf{135.3} & \textbf{23.2} & \textbf{73.3} & \textbf{43.1} & \textbf{160.1} & \textbf{76.01}\\
%\midrule
%\specialrule{0em}{1pt}{1pt}
\bottomrule
\end{tabular*}
\caption{The main results on VQA2.0 \cite{goyal2017making}, VizWizVQA \cite{gurari2018vizwiz}, SNLI\_VE \cite{xie2019visual}, COCOCaps \cite{lin2014microsoft}, TextCaps \cite{sidorov2020textcaps} and VizWizCaps \cite{gurari2020captioning} for full fine-tuning \cite{BERT,howard-ruder-2018-universal}, adapter \cite{sung2022vl}, prefix tuning \cite{Li2021PrefixTuningOC}, LoRA \cite{hu2021lora}, and our proposed $p$-adapter. We bold the scores for full fine-tuning and the highest scores separately for approaches with PETL methods.}
\label{table: main}
\end{table*}

We compare our method with full fine-tuning and other PETL methods, i.e., adapter \cite{houlsby2019parameter,sung2022vl}, prefix tuning \cite{Li2021PrefixTuningOC} and LoRA \cite{hu2021lora}. The results are shown in \Cref{table: main}. 

\minisection{Comparison with full fine-tuning.} In general, $p$-adapter is able to achieve comparable and even better performance than full fine-tuning on most benchmarks. Specifically, on VQA2.0 \cite{goyal2017making}, $p$-adapter outperforms full fine-tuning with mPLUG$_\text{ViT-B}$ while achieving comparable performance with BLIP$_\text{CapFilt-L}$. For VE task, $p$-adapter achieves improvements of 1.05\% and 0.54\% with BLIP$_\text{CapFilt-L}$ and mPLUG$_\text{ViT-B}$ on SNLI\_VE \cite{xie2019visual}, respectively. For image captioning, $p$-adapter surpasses full fine-tuning on COCO Captions \cite{lin2014microsoft}. The above experimental results demonstrate the effectiveness with good parameter efficiency for $p$-adapter (only tuning 6.39\% parameters). 
% However, full fine-tuning performs much better on TextCaps \cite{sidorov2020textcaps}. We conjecture this is due to the large gap between the downstream data and the pre-training corpus, which requires more parameter tuning to adapt to the new tasks. 

\minisection{Comparison with other PETL methods.} Our proposed method, $p$-adapter, outperforms other PETL methods with both BLIP$_\text{CapFilt-L}$ and mPLUG$_\text{ViT-B}$. Compared with full fine-tuning, prefix tuning \cite{Li2021PrefixTuningOC} suffers from a significant performance drop on almost all the benchmarks. With the same number of tunable parameters, LoRA \cite{hu2021lora} performs better than prefix tuning \cite{Li2021PrefixTuningOC} on all almost all the benchmarks. Adapter \cite{houlsby2019parameter,sung2022vl} outperforms these two methods with two times more tunable parameters. With only several extra trainable parameters compared with adapter \cite{houlsby2019parameter,sung2022vl} (the learnable $p$), $p$-Adapter achieves consistent improvements \cite{houlsby2019parameter,sung2022vl} and significantly outperforms all the PETL methods on all the benchmarks with the two pre-trained VLMs. Especially for VizWizVQA \cite{gurari2018vizwiz}, TextCaps \cite{sidorov2020textcaps} and VizWizCaps \cite{gurari2020captioning}, $p$-Adapter surpasses vanilla adapter with a large margin.
This demonstrates the effectiveness of our proposed attention re-normalization and feature aggregation  mechanisms inspired $p$-Laplacian message passing \cite{pmlr-v162-fu22e}.  

\subsection{Ablation Studies}

% \begin{table*}[]
% \centering
% \small
% \setlength\tabcolsep{4pt}
% \begin{tabular}{lcccccccccc}
% \toprule
% \multirow{3}{*}{Method} & VQA2.0 & VizWizVQA & SNLI\_VE & \multicolumn{2}{c}{COCOCaps} & \multicolumn{2}{c}{TextCaps} & \multicolumn{2}{c}{VizWizCaps} \\
%  & Karpathy test & test-dev & test-P & \multicolumn{2}{c}{Karpathy test} & \multicolumn{2}{c}{test-dev} & \multicolumn{2}{c}{test-dev} \\
%  & Acc.(\%) & Acc.(\%) & Acc.(\%) & BLEU@4 & CIDEr & BLEU@4 & CIDEr & BLEU@4 & CIDEr \\
% \midrule
% (A) Fixed $p$ \\
% (A-1) $p$=1.25 & 69.78 & 36.03 & 79.11 & 38.8 & 128.9 & 25.4 & 87.0 & 43.2 & 161.5 \\
% (A-2) $p$=1.50 & 69.76 & 36.02 & 78.91 & 38.7 & 128.6 & 25.6 & 86.9 & 43.2 & 161.3 \\
% (A-3) $p$=1.75 & 69.70 & 36.28 & 78.84 & 38.4 & 128.0 & 25.4 & 86.5 & 43.2 & 161.1 \\
% \midrule
% (B) All-Layer Learnable $p$ & \textbf{69.96} & & 79.12 & 38.8 & 128.7 & \textbf{25.9} & 87.0 & \textbf{43.5} & \textbf{162.0} \\
% (C) Layer-Wise Learnable $p$ & \textbf{69.96} & \textbf{36.58} & \textbf{79.14} & \textbf{39.0} &  \textbf{129.3} & 25.7 & \textbf{87.5} & 43.3 & 161.7 \\
% \bottomrule
% \end{tabular}
% \caption{Ablations on learnable $p$ strategy of the $p$-Adapter. We bold the highest scores for BLIP$_\text{CapFilt-L}$. }
% \label{table:ablation_p}
% \end{table*}

\begin{table}[tb!]
    \centering
    % \footnotesize
    % \setlength\tabcolsep{4pt}
    \resizebox{\linewidth}{!}{
        \begin{tabular}{lccccc}
            \toprule
            \multirow{2}{*}{GNN} & VQA2.0 & SNLI\_VE & \multicolumn{2}{c}{COCOCaps} & \multirow{2}{*}{Avg.} \\
            & Acc.(\%) & Acc.(\%) & BLEU@4 & CIDEr \\
            \midrule
            \midrule
            GCN & 69.53 & 78.85 & 38.9 & 128.8 & 79.02\\
            APPNP  & 70.22 & 79.03 & 39.4 & 129.1 & 79.44\\
            GCNII  & 70.13 & 79.12 & 39.7 & 129.7 & 79.66\\
            \midrule
            $^{p}$GNN (Ours)  & \textbf{70.39} & \textbf{79.40} & \textbf{40.4} &  \textbf{130.9} & \textbf{80.27}\\
            \bottomrule
        \end{tabular}
    }
    \caption{Ablation study on the graph neural networks. }
    \label{table:gnn-ablation}
\end{table}

\begin{table}[tb!]
    \centering
    % \small
    % \setlength\tabcolsep{4pt}
    \resizebox{\linewidth}{!}{
        \begin{tabular}{lccccc}
            \toprule
            \multirow{2}{*}{Concat.} & VQA2.0 & SNLI\_VE & \multicolumn{2}{c}{COCOCaps} & \multirow{2}{*}{Avg.} \\
            & Acc.(\%) & Acc.(\%) & BLEU@4 & CIDEr \\
            \midrule
            \midrule
            Zero  & 70.02 & 79.17 & 40.2 & 130.3 & 79.92\\
            Noise  & 69.90 & 78.99 & 39.9 & 130.1 & 79.72\\
            \midrule
            Query (Ours) & \textbf{70.39} & \textbf{79.40} & \textbf{40.4} &  \textbf{130.9} & \textbf{80.27} \\
            \bottomrule
        \end{tabular}}
    \caption{Ablation study on the concatenation pattern in augmented attention.}
    \label{table:ablation-concat}
\end{table}

\begin{table}[tb!]
    \centering
    % \small
    % \setlength\tabcolsep{4pt}
    \resizebox{\linewidth}{!}{
        \begin{tabular}{lccccc}
            \toprule
            \multirow{2}{*}{$p$ Values} & VQA2.0 & SNLI\_VE & \multicolumn{2}{c}{COCOCaps} & \multirow{2}{*}{Avg.}  \\
            & Acc.(\%) & Acc.(\%) & BLEU@4 & CIDEr \\
            \midrule
            \midrule
            Fixed $p=\,$1.25 & 70.38 & 78.84 & 40.3 & 130.8 & 80.08\\
            Fixed $p=\,$1.50 & 70.15 & 78.90 & 40.3 & 130.8 & 80.03\\
            Fixed $p=\,$1.75 & 70.34 & 78.94 & 40.1 & 130.7 & 80.02\\
            \midrule
            Learnable $p$ (Ours) & \textbf{70.39} & \textbf{79.40} & \textbf{40.4} & \textbf{130.9} & \textbf{80.27} \\
            \bottomrule
        \end{tabular}}
    \caption{Ablation study on learnable $p$ strategy of $p$-adapters.}
    \label{table:ablation_p}
\end{table}

\begin{table*}[tb!]
    \centering
    \resizebox{0.768\linewidth}{!}
    {
        \begin{tabular}{ccccccccc}
            \toprule
            \multirow{2}{*}{Method} & \multirow{2}{*}{FFN} & \multirow{2}{*}{SA} & \multirow{2}{*}{CA} & VQA2.0 & SNLI\_VE & \multicolumn{2}{c}{COCOCaps} & \multirow{2}{*}{Avg.}\\
            & & & & Acc. (\%) & Acc. (\%) & BLEU@4 & CIDEr \\
            \midrule
            \midrule
            \multirow{4}{*}{$p$-Adapter (Imps.)} & \Checkmark &  &  & 68.65 (-) & 78.21 (-) & 38.4 (-) & 128.4 (-) & 78.41 (-) \\
            & \Checkmark & \Checkmark &  & 70.11 (\textbf{+0.90}) & 78.96 (+0.34) & 39.9 (+1.4) & 130.3 (+1.8) & 79.82 (+1.11)\\
            & \Checkmark &  & \Checkmark & 69.84 (+0.67) & 79.17 (\textbf{+0.57}) & 39.1 (+0.5) & 129.4 (+0.7) & 79.38 (+0.61)\\
            & \Checkmark & \Checkmark & \Checkmark & \textbf{70.39} (+0.86) & \textbf{79.40} (+0.55) & \textbf{40.4} (\textbf{+1.5}) & \textbf{130.9} (\textbf{+2.1}) & \textbf{80.27} (\textbf{+1.25})\\
            \bottomrule
        \end{tabular}
    }
    \caption{Ablation study on different insertion positions, including FFN, self- and cross-attention. In the brackets, we show the improvements achieved by $p$-adapters over adapters. When only inserted after FFN, $p$-adapters and adapters are the same. We bold the best performance and the largest improvement for each task.}
    \label{table:ablation_adapter_pos}
\end{table*}

In the following section, we conduct five ablation studies to further illustrate the effectiveness of $p$-adapter. For the sake of our budgeted computation resources, all ablation studies are conducted with BLIP$_\text{CapFilt-L}$ on VQA2.0 \cite{goyal2017making}, SNLI\_VE \cite{xie2019visual} and COCO Captions \cite{lin2014microsoft}.

\minisection{Graph neural networks.} Within our modeling framework that casts adapter tuning to graph convolution, we test different GNNs for heterophilic graphs, namely APPNP \cite{gasteiger2018predict} and GCNII \cite{chen2020simple}. Note that GCN \cite{kipf2017semisupervised} equals to the vanilla adapter. The results are shown in \Cref{table:gnn-ablation}. Compared with vanilla adapters, adapters with message passing in APPNP~\cite{gasteiger2018predict} and GCNII~\cite{chen2020simple} achieve improvements on almost all the tasks, demonstrating the necessity of handling the heterophilic issues. Moreover, our proposed $p$-adapters further surpass these two designs. The major reason of adopting $p$-Laplacian message passing \cite{pmlr-v162-fu22e} is due to its better flexibility in handling graphs with different heterophily. \cite{pmlr-v162-fu22e} shows that different $p$-values can lead to different spectral properties, and our proposed learnable strategy for $p$ can dynamically adjust to attention graphs with various degrees of heterophily. 

\minisection{Concatenation in augmented attention.} In the augmented attention, we concatenate the query with the value features as the augmented value features, which are further used for re-normalization in $p$-Laplacian message passing. We test two more different ways of concatenation, i.e., padding random noise and zeros, for constructing the augmented value. The results are shown in \Cref{table:ablation-concat}. Concatenating the query vectors outperforms the other two alternatives, suggesting that introducing the query vectors leads to more information dipicting the attention mechanism, and thus necessitates the handing of the heterophilic attention graph.

\minisection{Learnable $p$ vs. fixed $p$.}
We compare two different strategies for determining $p$: fixed $p$ and learnable $p$. Specifically, we test three fixed different \textit{p} values \{1.25, 1.5, 1.75\}, and we try two learnable strategies, i.e., a unified learnable $p$ (all $p$-adapters share a $p$) and layer-wise learnable $p$. 
The results are shown in \Cref{table:ablation_p}.
According to \cite{pmlr-v162-fu22e}, different \textit{p} values provide different normalization intensity, thus leading to different spectral properties suitable for handling different graphs. As we can see, for fixed $p$, different tasks favor different $p$ values, which is consistent with the findings in \cite{pmlr-v162-fu22e}. 
Moreover, learnable strategies achieve substantial improvements compared with different fixed $p$ values.  

\minisection{Insertion positions.} 
We also test different insertion positions for adapters/$p$-adapters, including FFN, self-attention, and cross-attention. As shown in \Cref{table:ablation_adapter_pos}, for both adapters and $p$-adapters, more insertions (after both self-attention and cross-attention) lead to the best performance on all tasks. Moreover, the improvements from appending $p$-adapters/adapters after self/cross-attention differ in tasks. For instance, insertion after self-attention leads to more improvements for VQA2.0 \cite{goyal2017making} while SNLI\_VE \cite{xie2019visual} prefers insertion after cross-attention. Additionally, when only inserting after self-attention for VQA2.0 \cite{goyal2017making} and cross-attention for SNLI\_VE \cite{xie2019visual}, $p$-adapter achieves larger improvements than full insertion, which validates $p$-adapter's effectiveness in lower-resource scenario.

% \begin{table*}[tb!]
%     \centering
%     \resizebox{0.80\textwidth}{!}{
%         \begin{tabular}{cccccccc}
%             \toprule
%             \multirow{2}{*}{Method} & \multirow{2}{*}{\# of Layers} & Updated Params & VQA2.0 & SNLI\_VE & \multicolumn{2}{c}{COCOCaps} & \multirow{2}{*}{Avg.}\\
%             & & (\%) & Acc. (\%) & Acc. (\%) & BLEU@4 & CIDEr \\
%             \midrule
%             \multirow{3}{*}{$p$-Adapter (Imps.)} & 12 & 6.39 & \textbf{70.39} (+0.86) & \textbf{79.40} (+0.55) & \textbf{40.4} (+1.5) & \textbf{130.9} (\textbf{+2.1}) & \textbf{80.27} (+1.25) \\
%             & 9  & 4.79 & 69.42 (+0.29) & 78.60 (+0.44) & 38.9 (+0.4) & 128.6 (+0.5) & \\
%             & 6  & 3.19 & 68.44 (+0.64) & 78.18 (+0.31) & 38.8 (+0.1) & 128.8 (+0.5) & \\
%             & 3 & 1.59 & 66.04 (\textbf{+1.25}) & 77.15 (\textbf{+0.90}) & \textbf{39.1} (\textbf{+0.5}) & 128.8 (\textbf{+0.7}) & \\
%             \bottomrule
%         \end{tabular}
%         }
%         \caption{Ablation study on the number of transformer layers for insertion. In the bracket, we show the improvements achieved by $p$-adapters over adapters \cite{houlsby2019parameter}. We bold the best performance and the largest improvement for each task.}
%         \label{table:insertion-layers}
% \end{table*}

\minisection{Adapter size.}
\begin{figure}[tb!]
    \centering
    \vspace{-0.6cm}
    \subfloat[]{
        \label{fig:para-perf-vqa}
        \begin{tikzpicture}%
    \definecolor{Green-d}{RGB}{81,157,62}
    \definecolor{Green-l}{HTML}{d2e7ce}
    \definecolor{Blue-d}{HTML}{3a76af}
    \begin{axis}[
        scale only axis,clip=false,
        xmin=0, xmax=7.4,
        ymin=66.2, ymax=70.8,
        width =2.8cm,
        height=2.0cm,
        xtick={1, 2.5, 4, 5.5, 7},
        xticklabels={1, 2.5, 4, 5.5, 7},
        every major tick/.append style={%
            major tick length=2pt, black},
        x tick label style={font=\footnotesize},
        xlabel={Updated Params. (\%)},
        x label style={font=\footnotesize},
        ytick = {66.6, 67.9, 69.2, 70.5},
        yticklabels = {66.6, 67.9, 69.2, 70.5},
        y tick label style={font=\footnotesize},
        ylabel={VQA2.0 Acc. (\%)},
        y label style={at={(-0.2, 0.5)},font=\footnotesize},
        legend style={
            draw=none,fill=none,
            at={(1.08,1.0)},
            anchor=south east,
            legend columns=2,
            font=\footnotesize
        }
    ]
    \addplot[red, style={mark=o, mark size=1.0pt, line width=1.0pt, draw=Green-d}] table {data/vqa_adapter.dat};
    \addplot[blue, style={mark=o, mark size=1.0pt, line width=1.0pt, draw=Blue-d}] table {data/vqa_p_adapter.dat};
    \addplot[blue, style={mark=o, mark size=1.0pt, line width=1.0pt, draw=Blue-d}] table {data/vqa_p_adapter.dat};
    \legend{Adapter,$p$-Adapter}
    \draw[stealth-stealth,line width=1pt,gray!75] (0.79, 66.94)--(0.79, 69.69);
    \node[label={[label distance=-1mm]90:{\tiny\sffamily +2.75}}] at (axis cs:0.79, 69.69) {};
    \end{axis}%
\end{tikzpicture}%
    }
    \subfloat[]{
        \label{fig:para-perf-ve}
        \begin{tikzpicture}%
    \definecolor{Green-d}{RGB}{81,157,62}
    \definecolor{Blue-d}{HTML}{3a76af}
    \begin{axis}[
        scale only axis,clip=false,
        xmin=0, xmax=7.4,
        ymin=76.8, ymax=79.9,
        width =2.8cm,
        height=2.0cm,
        xtick={1, 2.5, 4, 5.5, 7},
        xticklabels={1, 2.5, 4, 5.5, 7},
        every major tick/.append style={%
            major tick length=2pt, black},
        x tick label style = {font=\footnotesize},
        xlabel={Updated Params.(\%)},
        x label style={font=\footnotesize},
        ytick = {76.4, 77.4, 78.4, 79.4},
        yticklabels = {76.4, 77.4, 78.4, 79.4},
        y tick label style = {font=\footnotesize},
        ylabel={SNLI\_VE Acc.(\%)},
        y label style={at={(-0.2, 0.5)}, font=\footnotesize},
        legend style={
            draw=none,fill=none,
            at={(1.08,1.0)},
            anchor=south east,
            legend columns=2,
            font=\footnotesize
        }
    ]
    \addplot[red, style={mark=o, mark size=1.0pt, line width=1.0pt, draw=Green-d}] table {data/snli_ve_adapter.dat};
    \addplot[blue, style={mark=o, mark size=1.0pt, line width=1.0pt, draw=Blue-d}] table {data/snli_ve_p_adapter.dat};
    \draw[stealth-stealth,line width=1pt,gray!75] (1.59,77.82)--(1.59,79.36);
    \node[label={[label distance=-.8mm]90:{\tiny\sffamily +1.54}}] at (axis cs:1.59,79.26) {};
    \legend{Adapter,$p$-Adapter}
    \end{axis}%
\end{tikzpicture}%
    }
    \caption{Ablation study on the adapter size. We report the results on VQA2.0 \cite{goyal2017making} and SNLI\_VE \cite{xie2019visual}.}
    \label{fig:para-perf}
\end{figure}
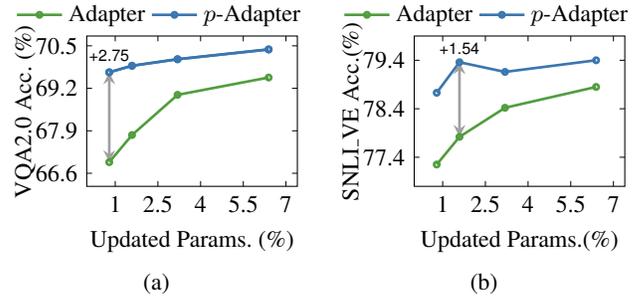
% \begin{figure*}[tb!] 
%     \centering
%     \includegraphics[width=1.00\linewidth]{figs/visualization.pdf}
%     \caption{Attention visualization on data from VQA2.0 \cite{goyal2017making}. We choose the cross-attention weights at the last transformer layer. Note that for $p$-adapter, we visualize the normalized attention matrix $\Bar{\vec{M}}$ in \Cref{eq:p-adapter-normalization}.} 
%     \label{fig:vis}
%     \vspace{-0.2cm}
% \end{figure*}
We also verify the performance with different adapter sizes. Specifically, we vary the hidden dimension (originally set to 96) of the learnable matrices in adapters, and the results on VQA2.0 \cite{goyal2017making} and SNLI\_VE \cite{xie2019visual} are shown in \Cref{fig:para-perf}. We can observe that the advantages of $p$-adapters are further enhanced with smaller adapters. For SNLI\_VE \cite{xie2019visual}, $p$-adapters outperform adapters with 1.54\%  when tuning 1.58\% parameters (the hidden dimension set to 24), and the improvement becomes 2.75\% when tuning 0.79\% parameters (the hidden dimension set to 12) for VQA2.0 \cite{goyal2017making}.

% \minisection{Number of inserted adapters.}
% We also investigate the impact of the number of transformer layers in which we insert adapters. We insert $p$-adapters/adapters in 3 (\nth{10}-\nth{12}), 6 (\nth{7}-\nth{12}), 9 (\nth{4}-\nth{12}), and 12 (\nth{1}-\nth{12}) layers of transformers. The results are shown in \Cref{table:insertion-layers}. We can observe that insertion in more layers leads to consistent improvements on most tasks for both $p$-adapters and adapters. Moreover, when inserting in fewer layers, the advantages for $p$-adapters over adapters are further enhanced. For instance, the improvements are 1.25\%, 0.9\%, 0.5 and 0.7 for the accuracy on VQA2.0 \cite{goyal2017making}, the accuracy on SNLI\_VE \cite{xie2019visual}, and BLEU@4 and CIDEr on COCO Captions \cite{lin2014microsoft}, respectively. This indicates that $p$-adapters have much better parameter efficiency than adapters. 

\subsection{Visualization}

To validate the effectiveness of $p$-adapter, we visualize \cite{chefer2021transformer} the cross-attention weights at the last transformer layer on some VQA \cite{goyal2017making} data, as shown in \Cref{fig:vis}. We take the  \texttt{[CLS]} token as the query since it represents the whole question and plot the attention weights on the image features in the key/value space. 
As we can see, with the normalized attention weights $\Bar{\vec{M}}$, $p$-adapter can dynamically calibrate the attention weights to focus on the relevant regions in the images that adapter ignores, which further improves the performance on downstream VL tasks. 
\section{Conclusion}
In this paper, we first propose a new modeling framework for adapter tuning \cite{houlsby2019parameter, sung2022vl} after attention modules in pre-trained VLMs. Specifically, we model it into a graph message passing process on attention graphs, where the nodes and edge weights are the projected query and value features, and the attention weights, respectively. Within this framework, we can identify the heterophilic nature of the attention graphs, posing challenges for vanilla adapter tuning \cite{houlsby2019parameter,sung2022vl}. To mitigate this issue, we propose a new adapter architecture, $p$-adapter, appended after the attention modules. Inspired by $p$-Laplacian message passing \cite{pmlr-v162-fu22e}, $p$-adapters re-normalize the attention weights using node features and aggregate the features with the calibrated attention matrix. Extensive experimental results validate our method's significant superiority over other PETL methods on various tasks, including VQA, VE and image captioning. 

% \minisection{Limitation.} The main limitation for $p$-adapter is the inference speed. Since we impose re-normalization and aggregation as shown in \Cref{eq:p-laplacian-forward} and \Cref{eq:p-adapter-normalization}, $p$-adapters take about 33\% more time than adapters for inference.   

\clearpage
{
\bibliography{aaai24.bbl}
}

\clearpage
\appendix
\setcounter{secnumdepth}{2} %May be changed to 1 or 2 if section numbers are desired.
\section{Appendix}
\subsection{Vision-Language Dataset Details}
\label{appendix:A}
We evaluate our models on six visual-language benchmarks: VQA2.0 \cite{goyal2017making} and VizWizVQA \cite{gurari2018vizwiz} for visual question answering (VQA), SNLI\_VE \cite{xie2019visual} for visual entailment (VE), and COCOCaps \cite{lin2014microsoft}, TextCaps \cite{sidorov2020textcaps}, and VizWizCaps \cite{gurari2020captioning} for image captioning. 
For VQA, instead of formulating VQA as a multi-answer classification task \cite{chen2020uniter}, we consider VQA as an answer generation problem, following \cite{cho2021unifying, li2022blip, li2021align}.
For VE, following \cite{song2022clip}, we first convert entailment, contradiction, and neutrality into correct, incorrect, and ambiguous as the answer list. Then, we prepend a prompt “Is it correct that” to the input text. Finally, the decoder generates the answer from the answer list. 
For image captioning, a prompt, ``What does the image describe?'', is prepended to the input text.
The statistics of each dataset are
shown in \Cref{table:dataset}. Note that we do not have the ground-truth labels for the test set of VizWizVQA \cite{gurari2018vizwiz}, VizWizCaps \cite{gurari2020captioning}, and TextCaps \cite{sidorov2020textcaps}, we thus split 10\% data out of the training set for hyper-parameter search and use the development set as the test set.

\begin{table}[!htbp]
  \scalebox{0.9}{
    \begin{tabular}{cccc}
      \toprule
      \multirow{2}{*}{Dataset} & \multicolumn{3}{c}{Data size (images/image-text pairs)} \\
      & Train & Dev & Test \\
      \midrule
      \multicolumn{1}{l}{VQA2.0} & 113.2K/605.1K & 5.0K/26.7K & 5.0K/26.3K \\
      \multicolumn{1}{l}{VizWizVQA} & 20.5K/205.2K & 4.3K/43.1K & 8.0K/80.0K \\
      \multicolumn{1}{l}{SNLI\_VE} & 29.8K/529.5K & 1K/17.8K & 1K/17.9K \\
      \multicolumn{1}{l}{COCOCaps} & 113.2K/566.8K & 5.0K/5.0K & 5.0K/5.0K \\
      \multicolumn{1}{l}{TextCaps} & 21.9K/109.7K & 3.1K/3.1K & 3.2K/3.2K \\
      \multicolumn{1}{l}{VizWizCaps} & 23.4K/117.1K & 7.7K/7.7K & 8.0K/8.0K \\
      \bottomrule
  \end{tabular}}
  \caption{The statistics of the datasets used in our experiments.}
  \label{table:dataset}
\end{table}

\subsection{Implementation Details}
\label{appendix:C}

We validate our method on two generative pre-trained vision-language models, BLIP$_\text{CapFilt-L}$ \cite{li2022blip} and mPLUG$_\text{ViT-B}$ \cite{li2022mplug}. BLIP$_\text{CapFilt-L}$ \cite{li2022blip} is constructed based on a multi-modal encoder-decoder architecture, and uses captioning and filtering strategy to learn from noisy image-text pairs during pre-training. Meanwhile, mPLUG$_\text{ViT-B}$ \cite{li2022mplug} is a vision-language model with novel cross-modal skip-connections. Instead of fusing visual and linguistic representations at the same levels, the cross-modal skip-connections enable the fusion to occur at disparate levels. 

Both BLIP$_\text{CapFilt-L}$ \cite{li2022blip} and mPLUG$_\text{ViT-B}$ \cite{li2022mplug} are encoder-decoder models, which have approximately twice the number of parameters as the decoder models. In order to improve the efficiency of transfer learning, following the training process of BLIP$_\text{CapFilt-L}$ \cite{li2022blip} on image captioning tasks, we reconstruct the original framework without using text encoder, and unify vision-language understanding and generation tasks through visual question answering format \cite{song2022clip}. Specifically, we use the encoder of BLIP$_\text{CapFilt-L}$ \cite{li2022blip} / mPLUG$_\text{ViT-B}$ \cite{li2022mplug} to encode the images, and the decoder of BLIP$_\text{ViT-B}$ \cite{li2022blip} / mPLUG$_\text{ViT-B}$ \cite{li2022mplug} to generate the answers in an auto-regressive way. 

\begin{table}[!htbp]
\centering
\resizebox{0.628\linewidth}{!}{
\begin{tabular}{lcc}
\toprule
Method                    & lr & epoch  \\
\midrule
(A) full fine-tuning      &  $2 \times 10^{-5} $ & 5      \\
\midrule
(B) prompt tuning          &     \\
prompt\_length=16 &  $1 \times 10^{-3} $ & 5       \\
\midrule
(C) lora                   &     \\
$\boldsymbol{r}$=8, $\alpha$=32    &  $1 \times 10^{-3} $ & 5      \\
\midrule
(D) adapter                &     \\
dim=\{96, 48, 24, 12\}              &  $2 \times 10^{-4} $ & 5      \\
\midrule
(E) fixed $p$-adapter        &     \\
dim=\{96, 48, 24, 12\}, &  \\
$p$=\{1.25, 1,5, 1,75\}, & \\
$\mu$=\{0.1, 1, 10\},       &  $5 \times 10^{-4} $ & 5  \\
\midrule
(F) learnable $p$-adapter    &     \\
dim=\{96, 48, 24, 12\}, &  \\
$p$=\{1.25, 1,5, 1,75\}, & \\
$\mu$=\{0.1, 1, 10\},       &  $5 \times 10^{-4} $ & 5  \\
\bottomrule
\end{tabular}}
\caption{VQA2.0 \cite{goyal2017making} and SNLI\_VE \cite{xie2019visual}}
\label{table: vqa2.0 and snli_ve}
\end{table}

\begin{table}[!htbp]
\centering
\resizebox{0.628\linewidth}{!}{
\begin{tabular}{lcc}
\toprule
Method                    & lr & epoch  \\
\midrule
(A) full fine-tuning      &  $2 \times 10^{-5} $ & 5    \\
\midrule
(B) prompt tuning         &  &     \\
prompt\_length=16   &  $5 \times 10^{-4} $ & 5    \\
\midrule
(C) lora                  &  &      \\
$\boldsymbol{r}$=8, $\alpha$=32    &  $5 \times 10^{-4} $ & 5    \\
\midrule
(D) adapter               &  &      \\
dim=96              &  $5 \times 10^{-4} $ & 5     \\
\midrule
(E) fixed $p$-adapter              \\
dim=96, &  \\
$p$=\{1.25, 1,5, 1,75\}, & \\
$\mu$=\{0.1, 1, 10\},       &  $5 \times 10^{-4} $ & 5  \\
\midrule
(F) learnable $p$-adapter         \\
dim=96, &  \\
$p$=\{1.25, 1,5, 1,75\}, & \\
$\mu$=\{0.1, 1, 10\},       &  $5 \times 10^{-4} $ & 5  \\
\bottomrule
\end{tabular}}
\caption{VizWizVQA \cite{gurari2018vizwiz}}
\label{table: vizwizvqa}
\end{table}

Both BLIP$_\text{CapFilt-L}$ \cite{li2022blip} and mPLUG$_\text{ViT-B}$ \cite{li2022mplug} adopt a BERT$_{base}$ \cite{devlin2018bert} architecture (12 layers of transformers, 12 attention heads, hidden size 768) for the decoder, and the visual transformer with ViT-B/16 \cite{dosovitskiy2020image, clip} (12 layers of transformers, 12 attention heads, hidden size 768) is used as the encoder. We use the AdamW \cite{loshchilov2017decoupled} optimizer with a weight decay of 0.05 and a linear learning rate schedule. During training, we take random image crops of resolution $224 \times 224$ as input, and also apply RandAugment \cite{cubuk2020randaugment} to improve the generalization of vision encoders. We show the hyper-parameters that we use for tuning on the downstream vision-language tasks separately in \Cref{table: vqa2.0 and snli_ve}, \Cref{table: vizwizvqa}, and \Cref{table: captions}.

\begin{table}[!htbp]
\centering
\resizebox{0.628\linewidth}{!}{
\begin{tabular}{lcc}
\toprule
Method                    & lr & epoch \\
\midrule
(A) full fine-tuning      &  $2 \times 10^{-5} $ & 2     \\
\midrule
(B) prompt tuning               \\
prompt\_length=16 &  $2 \times 10^{-4} $ & 2      \\
\midrule
(C) lora                        \\
$\boldsymbol{r}$=8, $\alpha$=32    &  $2 \times 10^{-4} $ & 2      \\
\midrule
(D) adapter                     \\
dim=96              &  $2 \times 10^{-4} $ & 2      \\
\midrule
(E) fixed $p$-adapter        &     \\
dim=\{96, 48, 24, 12\}, &  \\
$p$=\{1.25, 1,5, 1,75\}, & \\
$\mu$=\{0.1, 1, 10\},       &  $5 \times 10^{-4} $ & 2  \\
\midrule
(F) learnable $p$-adapter    &     \\
dim=\{96, 48, 24, 12\}, &  \\
$p$=\{1.25, 1,5, 1,75\}, & \\
$\mu$=\{0.1, 1, 10\},       &  $5 \times 10^{-4} $ & 2  \\
\bottomrule
\end{tabular}}
\caption{COCOCaps \cite{lin2014microsoft}, TextCaps \cite{sidorov2020textcaps}, and VizWizCaps \cite{gurari2020captioning}}
\label{table: captions}
\end{table}

% \section{Adapter/$p$-Adapter in Image Encoder}

% \begin{table}[!htbp]
%   \centering
%   \setlength\tabcolsep{4pt}
%   \resizebox{1\linewidth}{!}{
%     \begin{tabular}{ccccc}
%       \toprule
%       \multirow{2}{*}{Method} & VQA2.0 & SNLI\_VE & \multicolumn{2}{c}{COCOCaps}  \\
%       & Acc. (\%) & Acc. (\%) & BLEU@4 & CIDEr \\
%       \midrule
%       \multicolumn{1}{c}{Adapter} & 70.05 (+0.52) & 78.20 (-0.65) & 39.1 (+0.2) & 130.5 (+1.2)  \\
%       \multicolumn{1}{c}{$p$-Adapter} & 70.46 (+0.50) & 78.48 (-0.66) & 39.6 (+0.6) & 131.1 (+1.8)  \\
%       \bottomrule
%   \end{tabular}}
%   \caption{Comparison on adapter/$p$-adapter added in encoders. In the brackets, we show the improvements achieved by adding adapters/$p$-adapters in image encoder.}
%   \label{table:adapter_image_enc}
% \end{table}
% In this section we compare the performance of w.~or w/o adding adapter/$p$-adapter in the image encoder, as shown in \ref{table:adapter_image_enc}. The results show that there is no significant performance drop when appending adapter/$p$-adapter in the image encoder. Therefore, we only conduct all the PETL methods on the text decoder, same as \cite{sung2022vl}.

\subsection{$p$-Laplacian message passing}
\label{appendix:C}
In this section, we briefly summarize the $p$-Laplacian foundations and message passing proposed in~\cite{pmlr-v162-fu22e} and~\cite{zhou2005regularization}. We consider an \emph{undirected} graph $\mathcal{G}=(\mathcal{V},\mathcal{E})$ where $\mathcal{V}$ is the node set with cardinality $N$, $\mathcal{E}\subseteq\mathcal{V}\times\mathcal{V}$ is the edge set. Let $\mathcal{H}(V)$ be the Hilbert space of all real-valued functions defined on the node set $\mathcal{\mathcal{V}}$ endowed with the well-defined inner product $\langle f,g\rangle_{\mathcal{H}(\mathcal{V})}=\sum_{v\in\mathcal{V}}f(v)g(v)$. The Hilbert space $\mathcal{H}_{\mathcal{E}}$ is similarly defined for the edge set $\mathcal{E}$. Let $u\sim v$ denote all neighbors of $v$ for simplicity.
\begin{definition}[Graph Gradient]
  \label{def:graph-grad}
  The graph gradient~\cite{zhou2005regularization} of the function $f\in\mathcal{H}(\mathcal{V})$ is a function in $\mathcal{H}(\mathcal{E})$. For any edge $e_{u,v}=[u,v]\in\mathcal{E}$, the function value of graph gradient $\nabla f$ is defined by
  \begin{equation}
    \nabla f(e_{u,v})=\sqrt{\frac{w(e_{u,v})}{d(v)}}f(v)-\sqrt{\frac{w(e_{u,v})}{d(u)}}f(u),
  \end{equation}
  where $w(e_{u,v})$ is the weight of edge $e_{u,v}$ and $d(v)=\sum_{u\sim v}d(u)$ is the degree of $v\in\mathcal{V}$.
\end{definition}

To establish the foundation of $p$-Laplacian, we still require the definition of graph divergence.
\begin{definition}[Graph Divergence]
  \label{def:graph-div}
  The graph divergence \cite{zhou2005regularization} is an operator $\divop:\mathcal{H}(\mathcal{E})\rightarrow\mathcal{H}(\mathcal{V})$ defined by
  \begin{equation}
    \divop{g}(v)=\sum_{u\sim v}\sqrt{\frac{w(e_{u,v})}{d(v)}}\left(g(e_{v,u})-g(e_{u,v})\right),
  \end{equation}
  for any $v\in\mathcal{V}$.
\end{definition}
The definition of $p$-Laplacian can be extended to the discrete analogue given the above definitions.
\begin{definition}[Graph $p$-Laplacian]
  The graph $p$-Laplacian \cite{zhou2005regularization,pmlr-v162-fu22e} is an operator $\Delta_p:\mathcal{H}(\mathcal{V})\rightarrow\mathcal{H}(\mathcal{V})$ defined by 
  \begin{equation}
    \Delta_p{f}=-\frac{1}{2}\divop\left(\|\nabla f\|^{p-2}\nabla f\right),
  \end{equation}
  for some appropriate $p\geq1$. With the help of~\Cref{def:graph-grad} and~\Cref{def:graph-div}, we can write the $p$-Laplacian in a discrete form~\cite{zhou2005regularization}
  \begin{multline}
    \Delta_pf(v)=\frac{1}{2}\sum_{u\sim v}\left(\|\nabla_uf\|^{p-2}+\|\nabla_vf\|^{p-2}\right)\cdot\\
    \frac{w(e_{u,v})}{\sqrt{d(v)}}\left(\frac{f(v)}{\sqrt{d(v)}}-\frac{f(u)}{\sqrt{d(u)}}\right),
    \label{eq:def-p-laplacian}
  \end{multline}
  where the norm of the graph gradient $\nabla f$ at vertex $v$ is defined by $\|\nabla_vf\|^2=\sum_{u\sim v}(\nabla f)^2(e_{u,v})$.
\end{definition}

Note that $p$ is a hyperparameter. Specially,~\Cref{eq:def-p-laplacian} degrades into the conventional normalized graph Laplacian $\vec{I}-\vec{D}^{-\frac{1}{2}}\vec{W}\vec{D}^{-\frac{1}{2}}$ where $\vec{W}$ and $\vec{D}$ are the adjacency matrix and \emph{diagonal} degree matrix, respectively. The definition of norm $\|\nabla f\|$ is slightly different in~\cite{pmlr-v162-fu22e} where they treat the power of graph norm $\|\cdot\|^{p-2}$ element-wise and re-formulate~\Cref{eq:def-p-laplacian} as
\begin{multline}
  \Delta_pf(v)=\sum_{u\sim v}\|\nabla{f}(e_{u,v})\|^{p-2}\cdot\\
  \frac{w(e_{u,v})}{\sqrt{d(v)}}\left(\frac{f(v)}{\sqrt{d(v)}}-
  \frac{f(u)}{\sqrt{d(u)}}\right).
  \label{eq:def-p-laplacian-re}
\end{multline}
\Cref{eq:def-p-laplacian-re} can be easily extended to vector-valued functions $f$, where the operations are treated element-wise~\cite{pmlr-v162-fu22e}.
\begin{definition}[Graph Function Variation]
  The variation of graph function \cite{pmlr-v162-fu22e} $f\in\mathcal{H}(\mathcal{V})$ over the graph $\mathcal{G}=(\mathcal{V}, \mathcal{E})$ is defined by
  \begin{equation}
    \mathcal{S}_p(f)=\frac{1}{2}\sum_{e\in\mathcal{E}}\left\|\nabla f(e)\right\|^p,
    \label{eq:variation-def}
  \end{equation}
  where $\|\cdot\|$ is the norm defined in the function domain of $f$.
\end{definition}
Note that the summation inside variation $\mathcal{S}_p(f)$ in~\eqref{eq:variation-def} is taken over the entire edge set $\mathcal{E}$, while the \emph{$p$-Dirichlet form} $\mathcal{S}_p(f)=\frac{1}{2}\sum_{v\in\mathcal{V}}\|\nabla_vf\|^p$ in~\cite{zhou2005regularization} is taken over the whole node set $\mathcal{V}$.

Given node features $\vec{X}\in\mathbb{R}^{N\times d}$ and new features $\vec{F}\in\mathbb{R}^{N\times d}$ whose $i$-th rows $\vec{X}_{i,:},\vec{F}_{i,:}$ represent corresponding feature vectors, consider function $f:\mathcal{V}\rightarrow\mathbb{R}^d$ such that $f(v_i)=\vec{F}_{i,:}$ where $v_i\in\mathcal{V}$ is the $i$-th node of $\mathcal{G}$. For convenience, we use the graph gradient tensor $\nabla\vec{F}\in\mathbb{R}^{N\times N\times d}$ of feature matrix $\vec{F}$ to depict the graph gradient $\nabla{f}$, where
\begin{equation}
  \nabla\vec{F}_{i,j}=\nabla{f}(e_{v_i,v_j})=\sqrt{\frac{W_{i,j}}{D_{j,j}}}\vec{F}_{j,:}-\sqrt{\frac{W_{i,j}}{D_{i,i}}}\vec{F}_{i,:}.
\end{equation}
Note that if edge $[v_i,v_j]\notin\mathcal{E}$ indicating $W_{i,j}=0$, we naturally have $\nabla\vec{F}_{i,j}$ is a zero vector. The \emph{$p$-Laplacian regularization problem} ~\cite{pmlr-v162-fu22e} is defined by
\begin{equation}
  \min_{\vec{F}}\mathcal{L}_p(\vec{F}):=\mathcal{S}_p(\vec{F})+\mu\sum_{i=1}^N\|\vec{F}_{i,:}-\vec{X}_{i,:}\|_2^2,
  \label{opt:p-laplacian}
\end{equation}
where $\mu>0$. Given the weight matrix $\vec{W}$, the variation $\mathcal{S}_p(\vec{F})=\sum_{[i,j]\in\mathcal{E}}\|\nabla\vec{F}_{i,j}\|^p$ is simply
\begin{equation}
  \mathcal{S}_p(f)=\frac{1}{2}\sum_{i=1}^N\sum_{j=1}^N\left\|\sqrt{\frac{W_{i,j}}{D_{j,j}}}\vec{F}_{j,:}-\sqrt{\frac{W_{i,j}}{D_{i,i}}}\vec{F}_{i,:}\right\|^p
  \label{eq:variation-def-mat}
\end{equation}
according to~\Cref{eq:variation-def}, where the norm $\|\cdot\|$ in~\Cref{eq:variation-def-mat} is a vector norm in $\mathbb{R}^d$.

The solution to~\eqref{opt:p-laplacian} is the optimal $\vec{F}^*$ we desire. An iterative method to solve $\vec{F}^*$ for general $p\geq 1$ is proposed in~\cite{zhou2005regularization} and applied to graph neural networks in~\cite{pmlr-v162-fu22e} called {\bfseries $p$-Laplacian message passing} following the iterative propagation
\begin{equation}
  \vec{F}^{(k+1)}=\vec{\alpha}^{(k)}\vec{D}^{-\frac{1}{2}}\vec{M}^{(k)}\vec{D}^{-\frac{1}{2}}\vec{F}^{(k)}+\vec{\beta}^{(k)}\vec{X},
  \label{eq:p-laplacian-message-passing}
\end{equation}
where $k$ denotes the iteration number, $\vec{F}^{(k)}$ is the feature matrix at the $k$-th iteration, $\vec{M}^{(k)}$ is the normalized adjacency matrix with entries $M_{i,j}^{(k)}=W_{i,j}\|\nabla\vec{F}_{i,j}^{(k)}\|^{p-2}$, diagonal matrices $\vec{\alpha}^{(k)}$ and $\vec{\beta}^{(k)}$ are defined by
\begin{equation}
  \alpha_{i,i}^{(k)}=\left(\sum_{j=1}^N\frac{M_{i,j}^{(k)}}{D_{i,i}}+\frac{2\mu}{p}\right)^{-1},\;\;\beta_{i,i}^{(k)}=\frac{2\mu}{p}\alpha_{i,i}^{(k)}.
  \label{eq:def-alpha-beta}
\end{equation}
\Cref{eq:p-laplacian-message-passing} not only provides an iterative method to compute $\vec{F}^*$ but also illustrates propagation scheme in graph neural networks. The properties of $p$-Laplacian is well-studied in~\cite{pmlr-v162-fu22e}:
\begin{proposition}[Spectral Properties~\cite{pmlr-v162-fu22e}]
  \label{prop:spectral}
  Given a connected graph $\mathcal{G}=(\mathcal{V},\mathcal{E},\vec{W})$ with node embeddings $\vec{F}$ and the $p$-Laplacian $\Delta_p$ with its $p$-eigenvectors $\{\vec{u}^{(l)}\}_{l=0,\cdots,N-1}$ and the $p$-eigenvalues $\{\lambda^{(l)}\}_{l=0,\cdots,N-1}$. Let $g_p(\lambda_{i-1}):=\alpha_{i,i}\sum_jD_{i,i}^{-\frac{1}{2}}M_{i,j}D_{j,j}^{-\frac{1}{2}}$ for $i=1,\cdots,N$ be the filters defined on the spectral domain of $\Delta_p$, where $M_{i,j}=W_{i,j}\|\nabla f([v_i,v_j])\|^{p-2}$, $\nabla f([v_i,v_j])$ is the graph gradient of the edge between node $v_i$ and node $v_j$. $N_i$ denotes the number of edges connected to $v_i$, $N_{\min}=\min_{1\leq i\leq N}N_i$ and $(k,l^*)=\argmax_{j,l}\frac{|u_j^{(l)}|}{\sqrt{D_{l,l}}}$. Then we have
  \begin{enumerate}
  \item When $p=2$, $g_p(\lambda_{i-1})$ works as low-high-pass filters.
  \item When $p>2$, $g_p(\lambda_{i-1})$ works as low-high-pass filters on node $v_i$ if $\|\nabla_{v_i}f\|\leq2^{\frac{p-1}{p-2}}$ and low-pass filters on $v_i$ otherwise.
  \item When $1\leq p<2$, $g_p(\lambda_{i-1})$ works as low-pass filters on node $v_i$ if $\|\nabla_{v_i}f\|\leq2(2\sqrt{N_k})^{\frac{1}{p-2}}$ and low-high-pass filters on $v_i$ otherwise.
  \end{enumerate}
\end{proposition}
\Cref{prop:spectral} provides theoretical analysis of the $p$-Laplacian message passing. It is clear that $p$-Laplacian message passing with $1\leq p<2$ can deal with heterophilic graphs effectively.

In our $p$-Adapter, we apply a one-layer $p$-Laplacian message passing with a learnable $p$. Denote the node features by $\vec{X}$. $\vec{A}$ is the adjacency matrix (following the conventional notation), $\vec{D}$ is the diagonal degree matrix and $\bar{\vec{A}}$ is the normalized adjacency matrix after the graph gradient, with entries $\bar{A}_{i,j}=A_{i,j}\|\nabla\vec{X}_{i,j}\|^{p-2}$. The $p$-Laplacian message passing in~\Cref{eq:p-laplacian-message-passing} becomes
\begin{equation}
  \vec{X}'=\vec{\alpha}\vec{D}^{-\frac{1}{2}}\bar{\vec{A}}\vec{D}^{-\frac{1}{2}}\vec{X}+\vec{\beta}\vec{X},
\end{equation}
where $\vec{\alpha}$ and $\vec{\beta}$ are defined similarly as~\Cref{eq:def-alpha-beta}.

\end{document}